\documentclass[runningheads]{llncs}
\usepackage{graphicx}
\usepackage{comment}
\usepackage{multirow}
\usepackage{booktabs}
\usepackage{siunitx}
\usepackage{amsmath,amssymb} %
\usepackage{color}
\usepackage{wrapfig}
\usepackage{adjustbox}

\begin{document}
\pagestyle{headings}
\mainmatter
\def\ECCVSubNumber{1636}  %

\title{Spatio-Temporal Graph Transformer Networks for Pedestrian Trajectory Prediction} %

\titlerunning{Spatio-Temporal Graph Neural Networks}
\author{Cunjun Yu\thanks{equal contribution, listed in alphabetical order}\inst{1} \and
Xiao Ma$^\star$\inst{1,2}\and
Jiawei Ren\inst{1}\and
Haiyu Zhao\inst{1}\and
Shuai Yi\inst{1}
}
\authorrunning{Yu C., Ma X., Ren J., Zhao H., Yi S.}
\institute{SenseTime Research, \email{yucunjun@sensetime.com}
\and
National University of Singapore, \email{xiao-ma@comp.nus.edu.sg}
}

\maketitle

\begin{abstract}
Understanding crowd motion dynamics is critical to real-world applications, e.g., surveillance systems and autonomous driving. 
This is challenging because it requires effectively modeling the socially aware crowd spatial interaction and complex temporal dependencies. We believe attention is the most important factor for trajectory prediction. In this paper, we present \emph{STAR}, a Spatio-Temporal grAph tRansformer framework, which tackles trajectory prediction by only attention mechanisms. 
STAR models intra-graph crowd interaction by \emph{TGConv}, a novel Transformer-based graph convolution mechanism. The inter-graph temporal dependencies are modeled by separate temporal Transformers. STAR captures complex spatio-temporal interactions by interleaving between spatial and temporal Transformers. To calibrate the temporal prediction for the long-lasting effect of disappeared pedestrians, we introduce a read-writable external memory module, consistently being updated by the temporal Transformer.
We show that with only attention mechanism, STAR achieves the state-of-the-art performance on 5 commonly used real-world pedestrian prediction datasets.\footnote[1]{code available at https://github.com/Majiker/STAR}

\end{abstract}

\section{Introduction}

Crowd trajectory prediction is of fundamental importance to both the computer vision~\cite{alahi2016social,gupta2018social,zhang2019sr,huang2019stgat,ivanovic2019trajectron} and robotics~\cite{luo2018porca,luo2019gamma} community. This task is challenging because 1) human-human interactions are multi-modal and extremely hard to capture, e.g., strangers would avoid intimate contact with others, while fellows tend to walk in group~\cite{zhang2019sr}; 2)
the complex temporal prediction is coupled with the spatial human-human interaction, e.g., humans condition their motions on the history and future motion of their neighbors~\cite{huang2019stgat}.
\begin{figure}[t]
    \centering
        \includegraphics[width=\linewidth]{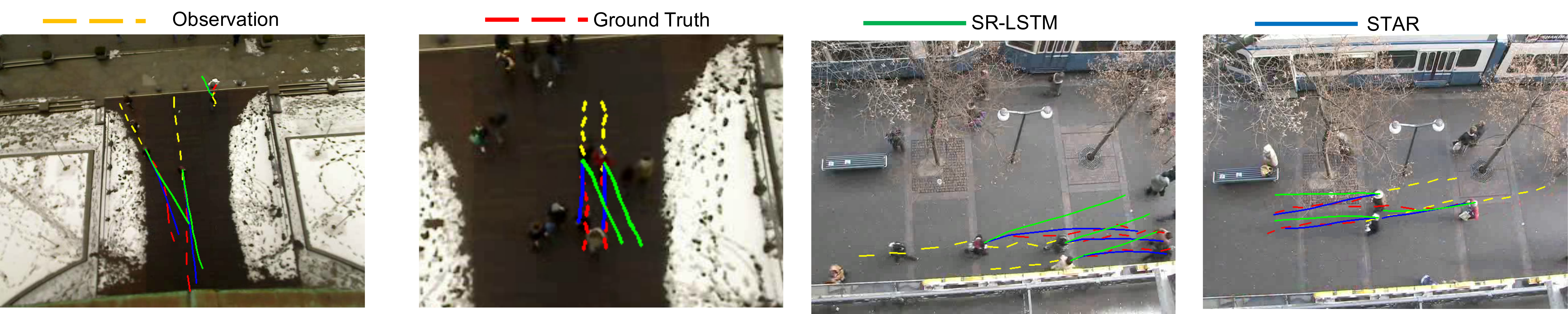}
        \caption{STAR successfully models spatio-temporal crowd dynamics with only a strong Transformer-based attention mechanism. STAR produces more accurate prediction trajectories compared to the state-of-the-art model, SR-LSTM.}
    
    \label{fig:overview}
\end{figure}
\begin{figure}[!t]
    \centering
    \begin{tabular}{c c}
        \includegraphics[height=60pt]{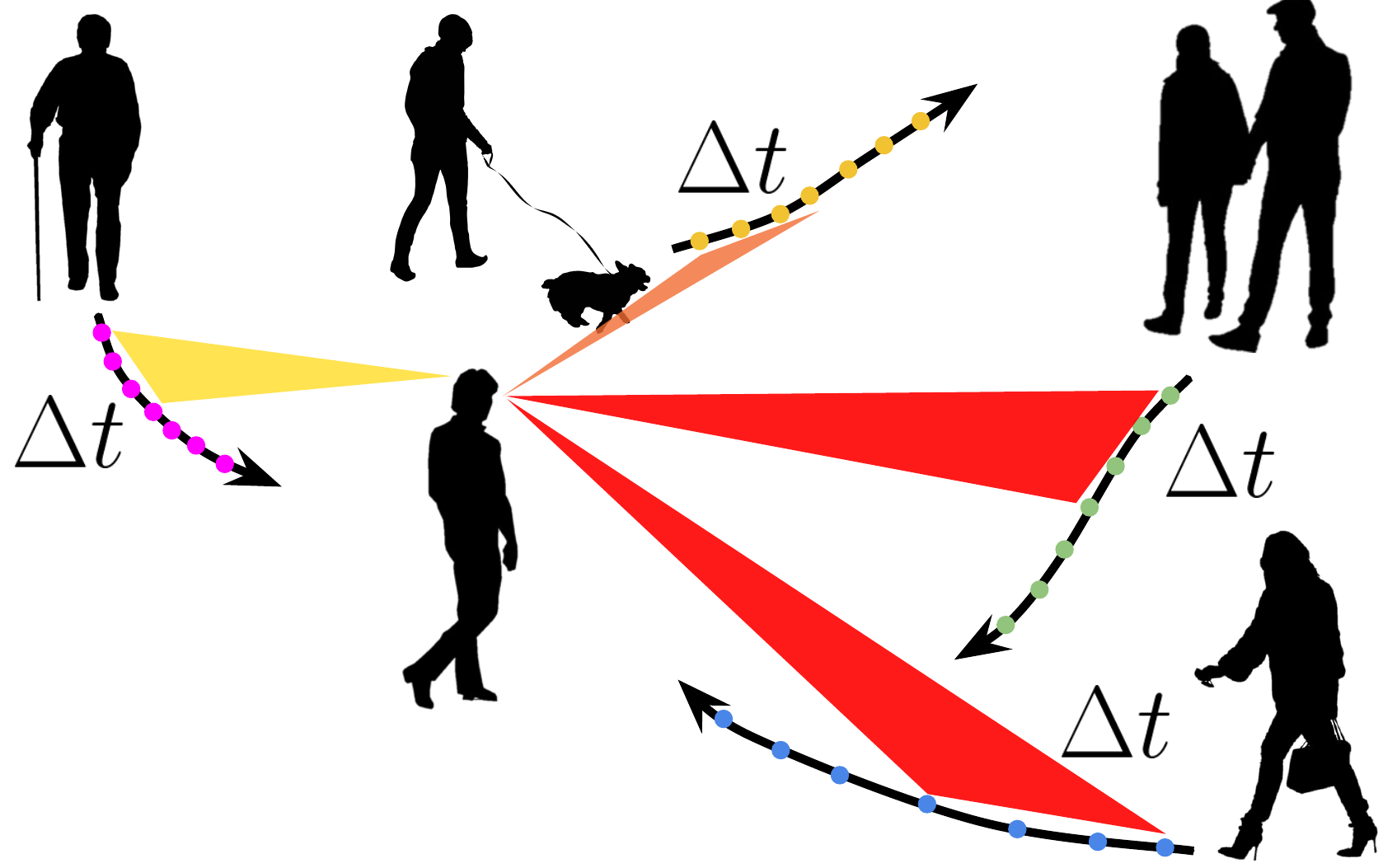}&
        \includegraphics[height=65pt]{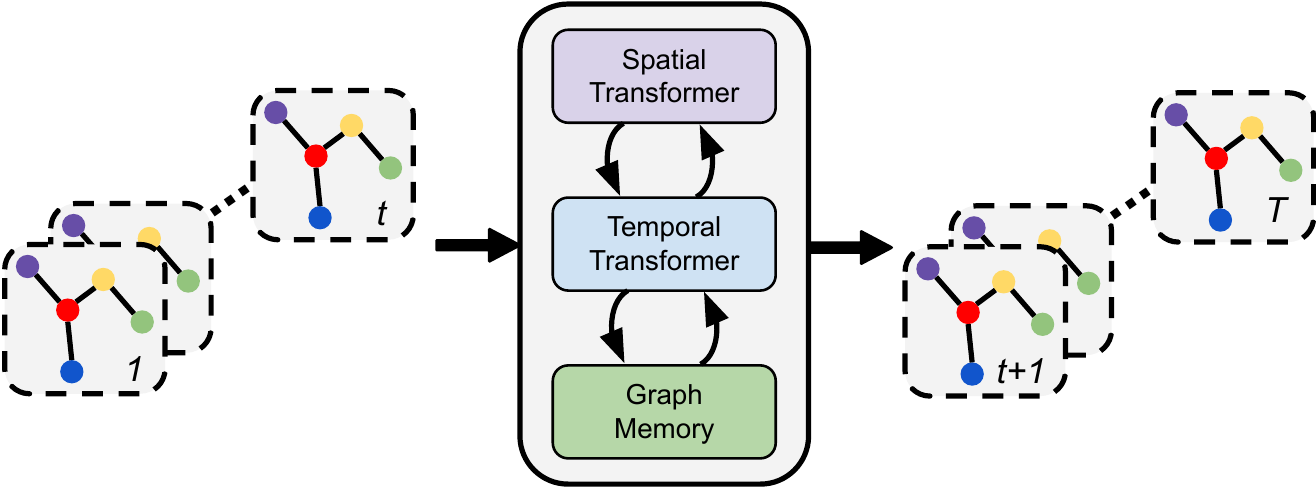}\\
         (a) Crowd Motion Modeling & (b) STAR Overview 
    \end{tabular}
    \centering
    \caption{ (a) People decide their future motions by paying different attentions (light yellow for less attention and dark red for more attention) to the potential future motions of their neighbors up to a certain time interval ($\Delta t$). (b) STAR models the crowd as a graph and learns spatio-temporal interaction of the crowd motion by interleaving between a graph-based spatial Transformer and a temporal Transformer. An external read-writable graph memory module is applied to improve the smoothness of the temporal predictions.
    }
    \label{fig:overview}
\end{figure}

Classic models capture human-human interaction by handcrafted energy-functions~\cite{helbing1995social,helbing2005self,luo2018porca}, which require significant feature engineering effort and normally fail to build crowd interactions in crowded spaces~\cite{huang2019stgat}. With the recent advances in deep neural networks, Recurrent Neural Networks (RNNs) have been extensively applied to trajectory prediction and demonstrated promising performance~\cite{alahi2016social,gupta2018social,zhang2019sr,huang2019stgat,ivanovic2019trajectron}. RNN-based methods capture pedestrian motion by their latent state and model the human-human interaction by merging latent states of spatially proximal pedestrians. Social-pooling~\cite{alahi2016social,gupta2018social} treat pedestrians in a neighborhood area equally and merge their latent state by a pooling mechanism. Attention mechanisms~\cite{ivanovic2019trajectron,zhang2019sr,huang2019stgat} relax this assumption and weigh pedestrians according to a learned function, which encodes unequal importance of neighboring pedestrians for trajectory prediction. However, existing predictors have two shared limitations: 1) the attention mechanisms used are still simple, which fails to fully model the human-human interaction, 2) RNNs normally have difficulty modeling complex temporal dependencies~\cite{vaswani2017attention}.

Recently, Transformer networks have made ground-breaking progress in Natural Language Processing domains (NLP)~\cite{vaswani2017attention,devlin2018bert,lan2019albert,young2018recent,yang2019xlnet}. Transformers discard the sequential nature of language sequences and model temporal dependencies with only the powerful self-attention mechanism. The major benefit of Transformer architecture is that self-attention significantly improves temporal modeling, especially for horizon sequences, compared to RNNs~\cite{vaswani2017attention}. Nevertheless, Transformer-based models are restricted to normal data sequences and it is hard to generalize them to more structured data, e.g., graph sequences. 

In this paper, we introduce the Spatio-Temporal grAph tRansformer (STAR) framework, a novel framework for spatio-temporal trajectory prediction based purely on self-attention mechanism. We believe that learning the temporal, spatial and temporal-spatial attentions is the key to accurate crowd trajectory prediction, and Transformers provide a neat and efficient solution to this task. STAR captures the human-human interaction with a novel spatial graph Transformer. In particular, we introduce \emph{TGConv}, a Transformer-based graph convolution mechanism. TGConv improves the attention-based graph convolution~\cite{velivckovic2017graph} by self-attention mechanism with Transformers and can capture more complex social interactions. Specifically, TGConv tends to improve more on datasets with higher pedestrian densities (ZARA1, ZARA2, UNIV). We model pedestrian motions with separate temporal Transformers, which better captures temporal dependencies compared to RNNs. STAR extracts spatio-temporal interaction among pedestrians by interleaving between spatial Transformer and temporal Transformer, a simple yet effective strategy. Besides, as Transformers treat a sequence as a bag of words, they normally have problem modeling time series data where strong temporal consistency is enforced~\cite{lim2019temporal}. We introduce an additional read-writable graph memory module that continuously performs smoothing over the embeddings during prediction.
An overview of STAR is given by Fig.~\ref{fig:overview}.(b)

We experimented on 5 commonly used real-world pedestrian trajectory prediction datasets. With only attention mechanism, STAR achieves the state-of-the-art on all 5 datasets. We conduct extensive ablation studies to better understand each proposed component.

\section{Background}
\subsection{Self-Attention and Transformer Networks}
Transformer networks have achieved great success in the NLP domain, such as machine translation, sentiment analysis, and text generation~\cite{devlin2018bert}. Transformer networks follow the famous encoder-decoder structure widely used in the RNN seq2seq models~\cite{bahdanau2014neural,cho2014learning}. 

The core idea of Transformer is to replace the recurrence completely by multi-head self-attention mechanism. For embeddings $\{h_t\}_{t=1}^T$, the self-attention of Transformers first learns the query matrix $Q=f_Q(\{h_t\}_{t=1}^T)$, key matrix $K=f_K(\{h_t\}_{t=1}^T)$ and a corresponding value matrix $V=f_V(\{h_t\}_{t=1}^T)$ of all embeddings from $t=1$ to $T$. It computes the attention by
\begin{equation}
    Att(Q, K, V) = \frac{\mbox{Softmax}(QK^{\mathrm{T}})}{\sqrt{d_k}}V\label{eq:att}
\end{equation}
where $d_k$ is the dimension of each query. The $1/\sqrt{d_k}$ implements the scaled-dot product term for numerical stability for attentions. By computing the self-attention between embeddings across different time steps, the self-attention mechanism is able to learn temporal dependencies over long time horizon, in contrast to RNNs that remember the history with a single vector with limited memory. Besides, decoupling attention into the query, key and value tuples allows the self-attention mechanism to capture more complex temporal dependencies.

Multi-head attention mechanism learns to combine multiple hypotheses when computing attentions. It allows the model to jointly attend to information from different representations at different positions. With $k$ heads, we have
\begin{align}
    \mbox{MultiHead}(Q, K, V) &= f_O(\left[\mbox{head}_i\right]_{i=1}^k)\nonumber\\
    \mbox{where head}_i &= Att_i(Q, K, V)\label{eq:mha}
\end{align}
where $f_O$ is a fully connected layer merging the output from $k$ heads and $Att_i(Q, K, V)$ denote the self-attention of the $i$-th head. Additional positional encoding is used to add positional information to the Transformer embeddings. Finally, Transformer outputs the updated embeddings by a fully connected layer with two skip connections. 

However, one major limitation of current Transformer-based models is they only apply to non-structured data sequences, e.g., word sequences. STAR extends Transformers to more structured data sequences, as a first step, graph sequences, and apply it to trajectory prediction.

\subsection{Related Works}
\subsubsection{Graph Neural Networks}
Graph Neural Networks (GNNs) are powerful deep learning architectures for graph-structured data. Graph convolutions~\cite{li2015gated,kipf2016semi,defferrard2016convolutional,gilmer2017neural,xu2018powerful} have demonstrated significant improvement on graph machine learning tasks, e.g., modeling physical systems~\cite{battaglia2016interaction,li2018learning}, drug prediction~\cite{liu2019chemi} and social recommendation systems~\cite{fan2019graph}. In particular, Graph Attention Networks (GAT)~\cite{velivckovic2017graph} implement efficient weighted message passing between nodes and achieved state-of-the-art results across multiple domains. From the sequence prediction perspective, temporal graph RNNs allow learning spatio-temporal relationship in graph sequences~\cite{cui2019traffic,hajiramezanali2019variational}. Our STAR improves GAT with TGConv, a transformer boosted attention mechanism and tackles the graph spatio-temporal modeling with transformer architecture.

\subsubsection{Sequence Prediction}
RNNs and its variants, e.g., LSTM~\cite{hochreiter1997long} and GRU~\cite{chung2014empirical}, have achieved great success in sequence prediction tasks, e.g., speech recognition~\cite{xiong2018microsoft,miao2015eesen}, robot localization~\cite{forster2007rnn,ma2019particle}, robot decision making~\cite{karkus2019differentiable,ma2020discriminative}, and etc. RNNs have been also successfully applied to model the temporal motion pattern of pedestrians~\cite{alahi2016social,gupta2018social,huang2019stgat,zhang2019sr,ivanovic2019trajectron}. RNNs-based predictors make predictions with a Seq2Seq structure~\cite{sutskever2014sequence}. Additional structure, e.g., social pooling~\cite{alahi2016social,gupta2018social}, attention mechanism~\cite{xu2018encoding,vemula2018social,ivanovic2019trajectron} and graph neural networks~\cite{huang2019stgat,zhang2019sr}, are used to improve the trajectory prediction with social interaction modeling.

Transformer networks have dominated Natural Language Processing domains in recent years~\cite{vaswani2017attention,devlin2018bert,lan2019albert,young2018recent,yang2019xlnet}. Transformer models completely discard the recurrence and focus on the attention across time steps. This architecture allows long-term dependency modeling and large-batch parallel training. Transformer architecture has also been applied to other domains with success, e.g., stock prediction~\cite{liu2019transformer}, robot decision making~\cite{fang2019scene} etc. STAR applies the idea of Transformer to the graph sequences. We demonstrate it on a challenging crowd trajectory prediction task, where we consider crowd interaction as a graph. STAR is a general framework and could be applied to other graph sequence prediction tasks, e.g., event prediction in social networks~\cite{ma2017beep} and physical system modeling~\cite{li2018learning}. We leave this for future study.

\subsubsection{Crowd Interaction Modeling}
As the pioneering work, Social Force models~\cite{helbing1995social,lohner2010modeling}, has been proven effective in various applications, e.g., crowd analysis~\cite{helbing2005self} and robotics~\cite{ferrer2013robot}. 
They assume the pedestrians are driven by virtual forces for goal navigation and collision avoidance. 
Social Force models work well on interaction modeling while performing poorly on trajectory prediction~\cite{kuderer2012feature}.
Geometry based methods, e.g., ORCA~\cite{van2011reciprocal} and PORCA~\cite{luo2018porca}, 
consider the geometry of the agent and convert the interaction modeling into an optimization problem.
One major limitation of classic approaches is that they rely on hand-crafted features, which is non-trivial to tune and hard to generalize.

Deep learning based models achieve automatic feature engineering by directly learning the model from data. Behavior CNNs~\cite{yi2016pedestrian} capture crowd interaction by CNNs. Social-Pooling~\cite{alahi2016social,gupta2018social} further encodes the proximal pedestrian states by a pooling mechanism that approximates the crowd interaction. Recent works consider crowd as a graph and merge information of the spatially proximal pedestrians with attention mechanisms~\cite{xu2018encoding,vemula2018social,ivanovic2019trajectron}. Attention mechanism models pedestrians with importance compared to the pooling methods. Graph neural networks are also applied to address crowd modeling~\cite{huang2019stgat,zhang2019sr}. Explicit message passing allows the network to model more complex social behaviors.

\section{Method}
\subsection{Overview}
In this section, we introduce the proposed spatio-temporal graph Transformer based trajectory prediction framework, STAR. We believe attention is the most important factor for effective and efficient trajectory prediction. 

STAR decomposes the spatio-temporal attention modeling into temporal modeling and spatial modeling. For temporal modeling, STAR considers each pedestrian independently and applies a standard temporal Transformer network to extract the temporal dependencies. The temporal Transformer provides a better temporal dependency modeling protocol compared to RNNs, which we validate in our ablation studies. 
For spatial modeling, we introduce \emph{TGConv}, a Transformer-based message passing graph convolution mechanism. TGConv improves the state-of-the-art graph convolution methods with a better attention mechanism and gives a better model for complex spatial interactions. In particular, TGConv tends to improve more on datasets with higher pedestrian densities (ZARA1, ZARA2, UNIV) and complex interactions. We construct two encoder modules, each including a pair of spatial and temporal Transformers, and stack them to extract spatio-temporal interactions. 

\subsection{Problem Setup}
We are interested in the problem of predicting future trajectories starting at time step $T_{obs}+1$ to $T$ of total $N$ pedestrians involved in a scene, given the observed history during time steps $1$ to $T_{obs}$. At each time step $t$, we have a set of $N$ pedestrians $\{p_t^i\}_{i=1}^{N}$, where $p_t^i=(x_t^i, y_t^i)$ denotes the position of the pedestrian in a top-down view map. We assume the pedestrian pairs $(p_t^i, p_t^j)$ with distance less than $d$ would have an undirected edge $(i, j)$. This leads to an \emph{interaction graph} at each time step $t$: $G_t = (V_t, E_t)$, where $V_t = \{p_t^i\}_{i=1}^{N}$ and $E_t = \{(i, j)\mid \mbox{$i, j$ is connected at time $t$}\}$. For each node $i$ at time $t$, we define its neighbor set as $Nb(i, t)$, where for each node $j\in Nb(i, t)$, $e_t(i, j)\in E_t$. 

\begin{figure}[!t]
    \centering
    \begin{tabular}{c c}
        \includegraphics[height=70pt]{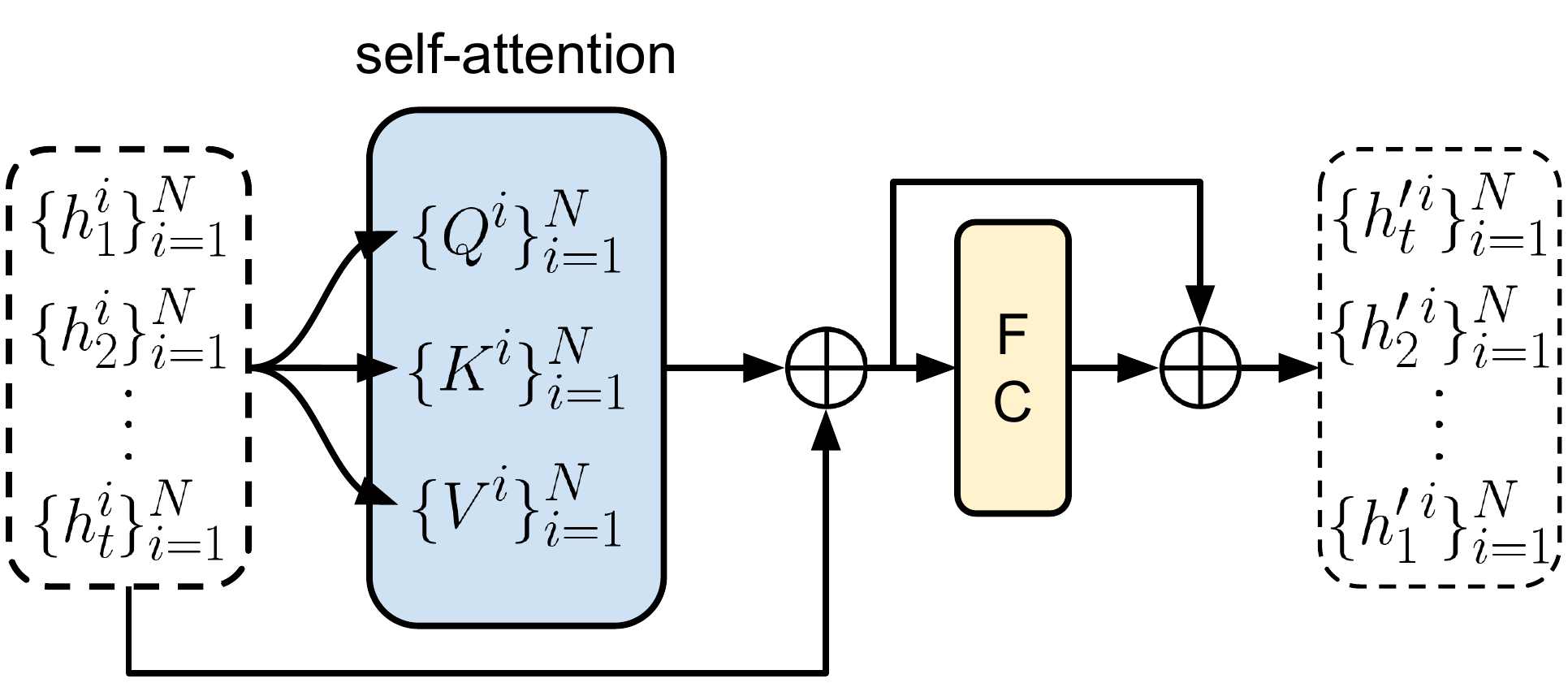}&
        \includegraphics[height=60pt]{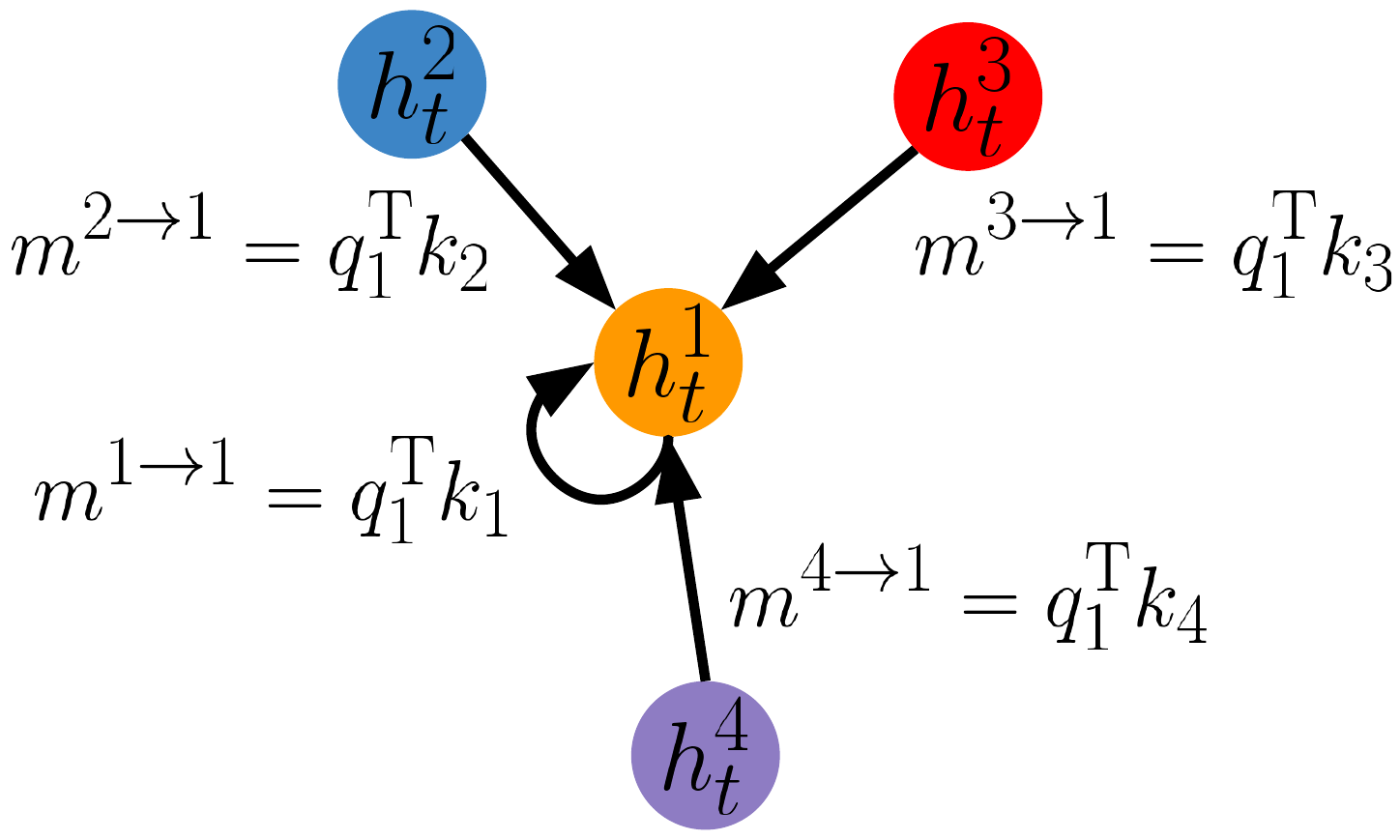}\\
         (a) Temporal Transformer & (b) Spatial Transformer
    \end{tabular}
    \centering
    \caption{ STAR has two main components, Temporal Transformer and Spatial Transformer. (a) Temporal Transformer treats each pedestrians independently and extracts the temporal dependencies by Transformer model ($h$ is the embedding of pedestrian positions, $Q$, $K$ and $V$ are the query, key, value matrix in Transformers). (b) Spatial Transformer models the crowd as a graph, and applies TGConv, a Transformer-based message passing graph convolution, to model the social interactions ($m^{i\rightarrow j}$ is the message from node $i$ to $j$ represented by Transformer attention)
    }
    \label{fig:transformers}
\end{figure}

\subsection{Temporal Transformer}
The temporal Transformer block in STAR uses a set of pedestrian trajectory embeddings $\{h_1^i\}_{i=1}^N, \{h_2^i\}_{i=1}^N, \dots, \{h_t^i\}_{i=1}^N$ as input, and output a set of updated embeddings $\{{h'}_1^i\}_{i=1}^N, \{{h'}_2^i\}_{i=1}^N, \dots, \{{h'}_t^i\}_{i=1}^N$ with temporal dependencies as output, considering each pedestrian independently. 

The structure of a temporal Transformer block is given by Fig.~\ref{fig:transformers}.(a). The self-attention block first learns the query matrices $\{Q^i\}_{i=1}^N$, key matrix $\{K^i\}_{i=1}^N$ and the value matrix $\{V^i\}_{i=1}^N$ given the inputs. For $i$-th pedestrian, we have
\begin{equation}
Q^i = f_Q(\{h_j^i\}_{j=1}^t), \quad K^i = f_K(\{h_j^i\}_{j=1}^t), \quad V^i = f_V(\{h_j^i\}_{j=1}^t)
\end{equation}
where $f_Q$, $f_K$ and $f_V$ are the corresponding query, key and value functions shared by pedestrians $i=1, \dots, N$. We could parallel the computation for all pedestrians, benefiting from the GPU acceleration. 

We compute the attention for each single pedestrian separately, following Eq.~\ref{eq:att}. Similarly, we have the multi-head attention ($k$ heads) for pedestrian $i$ represented as
\begin{align}
    Att(Q^i, K^i, V^i) &= \frac{\mbox{Softmax}(Q^iK^{i\mathrm{T}})}{\sqrt{d_k}}V^i\\
    \mbox{MultiHead}(Q^i, K^i, V^i) &= f_O (\left[head_j\right]_{j=1}^k)\\
    \mbox{where head}_j &= Att_j(Q^i, K^i, V^i)\label{eq:mha}
\end{align}{}
where $f_O$ is a fully connected layer that merges the $k$ heads and $Att_j$ indexes the $j$-th head. The final embedding is generated by two skip connections and a final fully connected layers, as shown in Fig.~\ref{fig:transformers}.(a). 

The temporal Transformer is a simple generalization of Transformer networks to a data sequence set. We demonstrate in our experiment that Transformer based architecture provides better temporal modeling. 

\subsection{Spatial Transformer}
The spatial Transformer block extracts the spatial interaction among pedestrians. We propose a novel Transformer based graph convolution, TGConv, for message passing on a graph.

Our key observation is that the self-attention mechanism can be regarded as message passing on an undirected fully connected graph. For a feature vector $h_i$ of feature set $\{h_i\}_{i=1}^n$, we can represent its corresponding query vector as $q_i = f_Q(h_i)$, key vector as $k_i = f_K(h_i)$ and value vector as $v_i = f_V(h_i)$. We define the message from node $j$ to $i$ in the fully connected graph as
\begin{equation}
m^{j\rightarrow i} = q_i^\mathrm{T}k_j
\end{equation}
and the attention function (Eq.~\ref{eq:att}) can be rewritten as
\begin{equation}
Att(Q, K, V) = \frac{\mbox{Softmax}\left(\left[m^{j\rightarrow i}\right]_{i,j=1:n}\right)}{\sqrt{d_k}}\left[v_i\right]_{i=1}^n
\end{equation}

Built upon the above insight, we introduce \emph{Transformer-based Graph Convolution (TGConv)}. TGConv is essentially an attention-based graph convolution mechanism, similar to GATConv~\cite{velivckovic2017graph}, but with a better attention mechanism powered by Transformers. For an arbitrary graph $G = (V, E)$ where $V=\{1, 2, \dots, n\}$ is the node set and $E = \{(i, j)\mid \mbox{$i, j$ is connected}\}$. Assume each node $i$ is associated with an embedding $h_i$ and a neighbor set $Nb(i)$. The graph convolution operation for node $i$ is written as
\begin{align}
    Att(i) &= \frac{\mbox{Softmax}\left(\left[m^{j\rightarrow i}\right]_{j\in Nb(i)\bigcup \{i\}}\right)}{\sqrt{d_k}}\left[v_j\right]^\mathrm{T}_{j\in Nb(i)\bigcup \{i\}} + h_i\\
    h'_i &= f_{out}(Att(i)) + Att(i)
\end{align}
where $f_{out}$ is the output function, in our case, a fully connected layer, and $h'_i$ is the updated embedding of node $i$ by TGConv. We summarize the TGConv function for node $i$ by $TGConv(h_i)$.
In a Transformer structure, we would normally apply layer normalization~\cite{ba2016layer} after each skip connection in the above equations. We ignored them in the equations for a clean notation.

The spatial Transformer, as shown in Fig.~\ref{fig:transformers}.(b), can be easily implemented by the TGConv. A TGConv with shared weights is applied to each graph $G_t$ separately. We believe TGConv is general and can be applied to other tasks and we leave it for future study.

\begin{figure}[t]
    \centering
    \includegraphics[width=.9\linewidth]{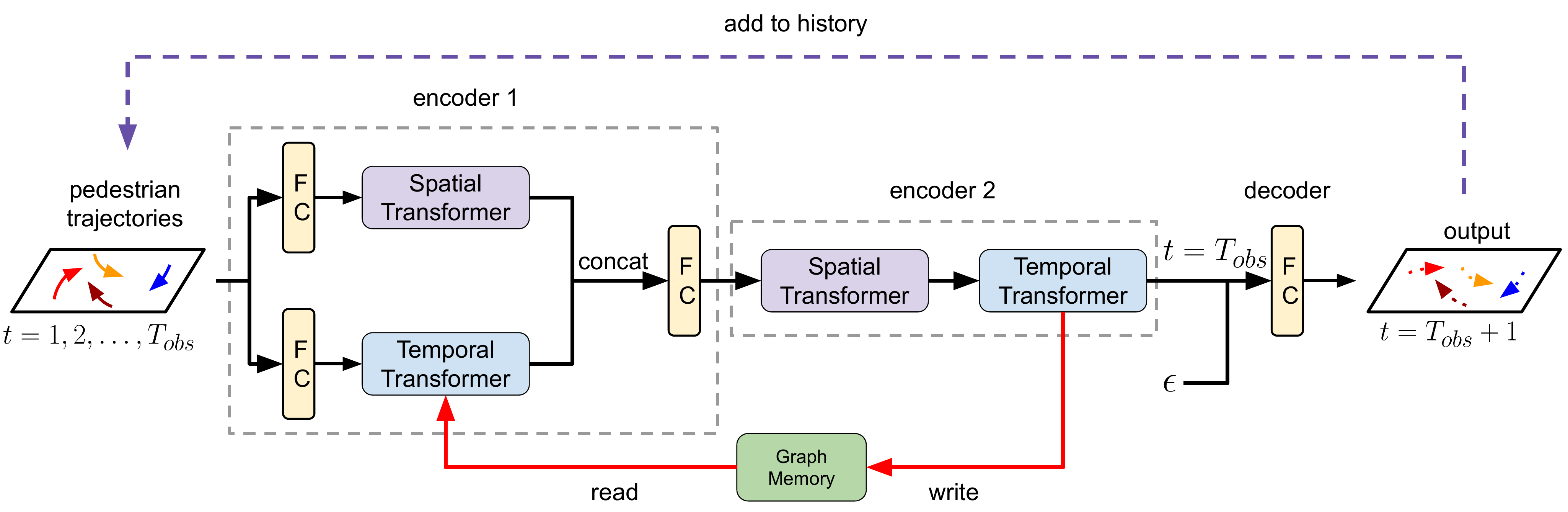}
    \caption{ Network structure of STAR with application to trajectory prediction. In STAR, trajectory prediction is achieved completely by attention mechanisms. STAR interleaves spatial Transformer and temporal Transformer in two encoder blocks to extract spatio-temporal pedestrian dependencies. An external read-writable graph memory module helps to smooth the graph embeddings and improve the consistency of temporal predictions. The prediction at $T_{obs} + 1$ is added back to history to predict the pedestrian poses at $T_{obs} + 2$.}
    \label{fig:st_grat} 
\end{figure}

\subsection{Spatio-Temporal Graph Transformer}
In this section, we introduce the Spatio-Temporal grAph tRansformer (STAR) framework for pedestrian trajectory prediction.

Temporal transformer can model the motion dynamics of each pedestrian separately, but fails to incorporate spatial interactions; spatial Transformer tackles crowd interaction with TGConv but can be hard to generalize to temporal sequences. One major challenge of pedestrian prediction is modeling coupled spatio-temporal interaction. The spatial and temporal dynamics of a pedestrian is tightly dependent on each other. For example, when one decides her next action, one would first predict the future motions of her neighbors, and choose an action that avoids collision with others in a time interval $\Delta t$. 

STAR addresses the coupled spatio-temporal modeling by interleaving the spatial and temporal Transformers in a single framework. Fig.~\ref{fig:st_grat} shows the network structure of STAR. STAR has two encoder modules and a simple decoder module. The input to the network is the pedestrian position sequences from $t=1$ to $t=T_{obs}$, where the pedestrian positions at time step $t$ is denoted by $\{p_t^i\}_{i=1}^N$ with $p_t^i = (x_t^i, y_t^i)$. In the first encoder, we embed the positions by two separate fully connected layers and pass the embeddings to spatial Transformer and Temporal Transformer, to extract independent spatial and temporal information from the pedestrian history. The spatial and temporal features are then merged by a fully connected layer, which gives a set of new features with spatio-temporal encodings. To further model spatio-temporal interaction in the feature space, we perform post-processing of the features with the second encoder module. In encoder 2, spatial Transformer models spatial interaction with temporal information; the temporal Transformer enhances the output spatial embeddings with temporal attentions. 
STAR predicts the pedestrians positions at $t=T_{obs} + 1$ using a simple fully connected layer with the $t=T_{obs}$ embeddings from the second temporal Transformer as input, concatenated with a random Gaussian noise to generate various future predictions~\cite{huang2019stgat}. We construct $G_{T_{obs} + 1}$ by connecting the nodes with distance smaller than $d$ according to the predicted positions. The prediction is added to the history for the next step prediction.

The STAR architecture significantly improves the spatio-temporal modeling ability compared to naively combining spatial and temporal Transformers.

\subsection{External Graph Memory}
Although Transformer networks improve long-horizon sequence modeling by self-attention mechanism, it would potentially have difficulties handling continuous time-series data which requires a strong temporal consistency~\cite{lim2019temporal}. 
Temporal consistency, however, is a strict requirement for trajectory prediction, because pedestrian positions normally would not change sharply during a short period. 

We introduce a simple external \emph{graph memory} to tackle this dilemma. A graph memory $M_{1:T}$ is read-writable and learnable, where $M_t(i)$ has the same size with $h_t^i$ and memorizes the embeddings of pedestrian $i$.
At time step $t$, in encoder 1, the temporal Transformer first reads from memory $M$ the past graph embeddings with function 
$\{\Tilde{h}_1^i, \Tilde{h}_2^i, \dots, \Tilde{h}_{t-1}^i\}_{i=1}^N = f_{read}(M)$ 
and concatenate it with the current graph embedding $\{h_t^i\}_{i=1}^N$. This allows the Temporal Transformers to condition current embeddings on the previous embedding for a consistent prediction. In encoder 2, we write the output 
$\{{h'}_1^i, {h'}_2^i, \dots, {h'}_{t}^i\}_{i=1}^N$ of Temporal Transformer to the graph memory by function $M'=f_{write}(\{{h'}_1^i, {h'}_2^i, \dots, {h'}_{t}^i\}_{i=1}^N, M)$, which performs a smoothing over the time series data. For any $t' < t$, the embeddings will be updated by the information from $t'' > t$,
which gives temporally smoother embeddings for a more consistent trajectory.

For implementing $f_{read}$ and $f_{write}$, many potential function forms could be adopted.
In this paper, we only consider a very simple strategy
\begin{gather}
\{\Tilde{h}_1^i, \Tilde{h}_2^i, \dots, \Tilde{h}_{t-1}^i\}_{i=1}^N = f_{read}(M) = \{M_1(i), M_2(i), \dots, M_{t-1}(i)\}_{i=1}^N\\
M'=f_{write}(\{{h'}_1^i, {h'}_2^i, \dots, {h'}_{t}^i\}_{i=1}^N, M) = \{{h'}_1^i, {h'}_2^i, \dots, {h'}_{t}^i\}_{i=1}^N
\end{gather}
that is, we directly replace the memory with the embeddings and copy the memory to generate the output. This simple strategy works well in practice. More complicated functional form of $f_{read}$ and $f_{write}$ could be considered, e.g., fully connected layers or RNNs. We leave this for future study.

\section{Experiments} 

In this section, we first report our results on five pedestrian trajectory datasets which serve as the major benchmark for the task of trajectory prediction: ETH (ETH and HOTEL) and UCY (ZARA1, ZARA2, and UNIV) datasets. We compare STAR to 9 trajectory predictors, including the SOTA model, SR-LSTM~\cite{zhang2019sr}. We follow the leave-one-out cross-validation evaluation strategy which is commonly adopted by previous works. We also perform extensive ablation studies to understand the effect of each proposed component and try to provide deeper insights for model design in the trajectory prediction task.

As a brief conclusion, we show that: 1) STAR outperforms the SOTA model on 4 out of 5 datasets and have a comparable performance to the SOTA model on the other dataset; 2) the spatial Transformer improves crowd interaction modeling compared to existing graph convolution methods; 3) the temporal Transformer generally improves the LSTM; 4) the graph memory gives a smoother temporal prediction and a better performance.

\subsection{Experiment Setup}
 We follow the same data prepossessing strategy as SR-LSTM\cite{zhang2019sr} for our method. The origin of all the input is shifted to the last observation frame. Random rotation is adopted for data augmentation.
 \begin{itemize}
    \item Average Displacement Error (ADE): the mean square error (MSE) overall estimated positions in the predicted trajectory and ground-truth trajectory.
    \item Final Displacement Error (FDE): the distance between the predicted final destination and the ground-truth final destination.
\end{itemize}
We take 8 frames (3.2s) as an sequence and 12 frames(4.8s) as the target sequence for prediction to have a fair comparison with all the existing works. 
\subsection{Implementation Details} 
Coordinates as input would be first encoded into a vector in size of 32 by a fully connected layer followed with ReLU activation. The dropout ratio at 0.1 is applied when processing the input data. All the transformer layers accept input with feature size at 32. Both spatial transformer and temporal transformer consists of encoding layers with 8 heads. We performed a hyper-parameter search over the learning rate, from 0.0001 to 0.004 with interval 0.0001 on a smaller network and choose the best-performed learning rate (0.0015) to train all the other models. As a result, we train the network using Adam optimizer with a learning rate of 0.0015 and batch size 16 for 300 epochs. Each batch contains around 256 pedestrians in different time windows indicated by an attention mask to accelerate the training and inference process.

\subsection{Baselines}
\label{baseline}
We compare STAR with a wide range of baselines, including: 1) LR: A simple temporal linear regressor; 2) LSTM: a vanilla temporal LSTM; 3) S-LSTM~\cite{alahi2016social}: each pedestrian is modeled with an LSTM, and the hidden state is pooled with neighbors at each time-step; 4) Social Attention~\cite{vemula2018social}: it models the crowd as a spatio-temporal graph and uses two LSTMs to capture spatial and temporal dynamics; 5) CIDNN~\cite{xu2018cidnn}: a modularized approach for spatio-temporal crowd trajectory prediction with LSTMs; 6) SGAN~\cite{gupta2018social}: a stochastic trajectory predictor with GANs; 7) SoPhie~\cite{Sadeghian2019sophie}: one of the SOTA stochastic trajectory predictors with LSTMs.  8) TrafficPredict~\cite{mayuexin2019trafficpredict}: LSTM-based motion predictor for heterogeneous traffic agents. Note that TrafficPredict in \cite{mayuexin2019trafficpredict} reports isometrically normalized results. We scale them back for a consistent comparison; 9) SR-LSTM: the SOTA trajectory predictor with motion gate and pair-wise attention to refine the hidden state encoded by LSTM to obtain social interactions.

\subsection{Quantitative Results and Analyses}
We compare STAR with state-of-the-art approaches as mentioned in  Section \ref{baseline}. 
All the stochastic method samples 20 times and reports the best-performed sample.

The main results are presented in Table~\ref{table:comparsion}. We observe that STAR-D outperforms SOTA deterministic models on the overall performance, and the stochastic STAR significantly outperforms all SOTA models by a large margin.

One interesting finding is that the simple model LR significantly outperforms many deep learning approaches including the SOTA model, SR-LSTM, in the HOTEL scene, which mostly contains straight-line trajectories and is relatively less crowded. This indicates that these complex models might overfit to those complex scenes like UNIV. Another example is that STAR significantly outperforms SR-LSTM on ETH and HOTEL, but is only comparable to SR-LSTM on UNIV, where the crowd density is high. This can potentially be explained by that SR-LSTM has a well-designed gated-structure for message passing on the graph, but has a relatively weak temporal model, a single LSTM. The design of SR-LSTM potentially improves spatial modeling but might also lead to overfitting. 
In contrast, our approach performs well in both simple and complex scenes. We then will further demonstrate this in Sect.~\ref{sect:qualitative} with visualized results.

\begin{table*}[h!]
\begin{center}
\begin{tabular}{c|c|c|c|c|c|c}
\hline %
 & \multicolumn{6}{c}{Performance (ADE/FDE)} \\
\hline
Deterministic & ETH  & HOTEL & ZARA1 & ZARA2 & UNIV &AVERAGE\\
\hline
\hline
LR &1.33/2.94 &0.39/0.72 &0.62/1.21 & 0.77/1.48 & 0.82/1.59 & 0.79/1.59\\
LSTM & 1.13/2.39 &0.69/1.47 & 0.64/1.43& 0.54/1.21 & 0.73/1.60 &0.75/1.62\\
S-LSTM\cite{alahi2016social}& 0.77/1.60 &0.38/0.80 & 0.51/1.19 & 0.39/0.89& 0.58/1.28 & 0.53/1.15\\
CIDNN\cite{xu2018cidnn}& 1.25/2.32 & 1.31/1.86 & 0.90/1.28 & 0.50/1.04 &\textbf{0.51}/\textbf{1.07} &0.89/1.73\\
SocialAttention \cite{vemula2018social}& 1.39/2.39 & 2.51/2.91 & 1.25/2.54 & 1.01/2.17 & 0.88/1.75 & 1.41/2.35 \\
TrafficPredict \cite{mayuexin2019trafficpredict}& 5.46/9.73 & 2.55/3.57 & 4.32/8.00 & 3.76/7.20 & 3.31/6.37 & 3.88/6.97\\
SR-LSTM \cite{zhang2019sr}&0.63/1.25 & 0.37/0.74 & \textbf{0.41}/\textbf{0.90} & 0.32/\textbf{0.70} &\textbf{0.51}/1.10 &0.45/0.94 \\
STAR-D & \textbf{0.56}/\textbf{1.11} & \textbf{0.26}/\textbf{0.50} & \textbf{0.41}/\textbf{0.90} &\textbf{0.31}/0.71 & 0.52/1.15 &\textbf{0.41}/\textbf{0.87}\\
\hline
Stochastic & ETH  & HOTEL & ZARA1 & ZARA2 & UNIV &AVERAGE\\
\hline
\hline
SGAN{$^\dagger$} \cite{gupta2018social}& 0.81/1.52 & 0.72/1.61 & 0.34/0.69 & 0.42/0.84 & 0.60/1.26 & 0.58/1.18 \\
SoPhie*{$^\dagger$} \cite{Sadeghian2019sophie}& 0.70/1.43 & 0.76/1.67 & 0.30/0.63 & 0.38/0.78 &  0.54/1.24 & 0.54/1.15 \\
STGAT{$^\dagger$}~\cite{huang2019stgat} &0.65/1.12 & 0.35/0.66    & 0.34/0.69 & 0.29/0.60  & 0.52/1.10 & 0.43/0.83\\
STAR{$^\dagger$} & \textbf{0.36}/\textbf{0.65} & \textbf{0.17}/\textbf{0.36}& \textbf{0.26}/\textbf{0.55}& \textbf{0.22}/\textbf{0.46}& \textbf{0.31}/\textbf{0.62}& \textbf{0.26}/\textbf{0.53}\\
\hline
\hline
\end{tabular}
\end{center}
\caption{Comparison with baselines models. STAR-D denotes the deterministic version of STAR. {$^\dagger$}: The results marked with {$^\dagger$} are calculated on 20 samples since they are stochastic models. {$*$}: SoPhie takes extra image input.}
\label{table:comparsion}
\end{table*}

\subsection{Qualitative Results and Analyses}\label{sect:qualitative}
\begin{figure}[t]
    \centering
    \includegraphics[width=\linewidth]{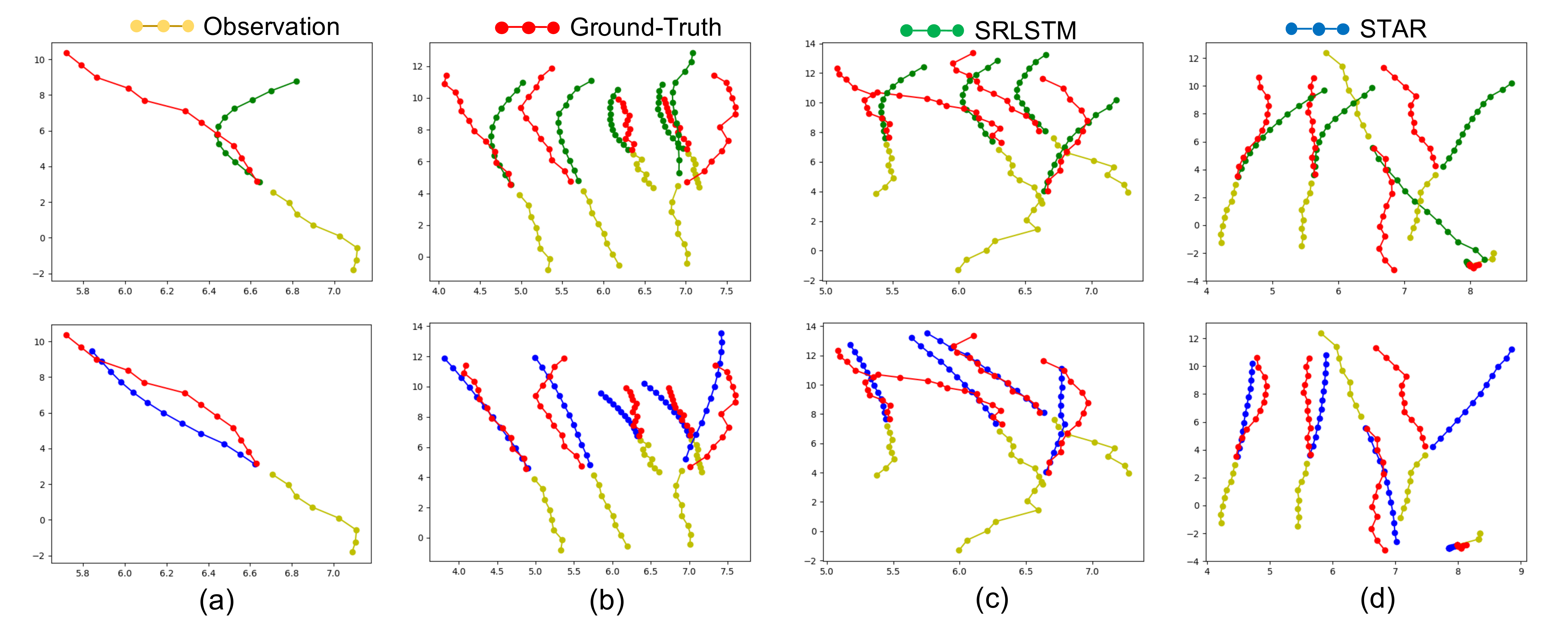}
    \caption{Trajectory visualization. STAR successfully models the spatio-temporal interaction of the crowd and makes better predictions than the SOTA model, SR-LSTM. (a) STAR accurately extracts the temporal dynamics of the agent; (b, c, d) STAR is able to model crowd interaction and spatio-temporal interactions. }
    \label{fig:dot_vis}
\end{figure}

We present our qualitative results in Fig.~\ref{fig:dot_vis} and Fig.~\ref{fig:att}.
\begin{figure*}[t]
\fontsize{6}{11}\selectfont
    \centering
    \begin{tabular}{c c c c}
    \includegraphics[width=0.22\linewidth]{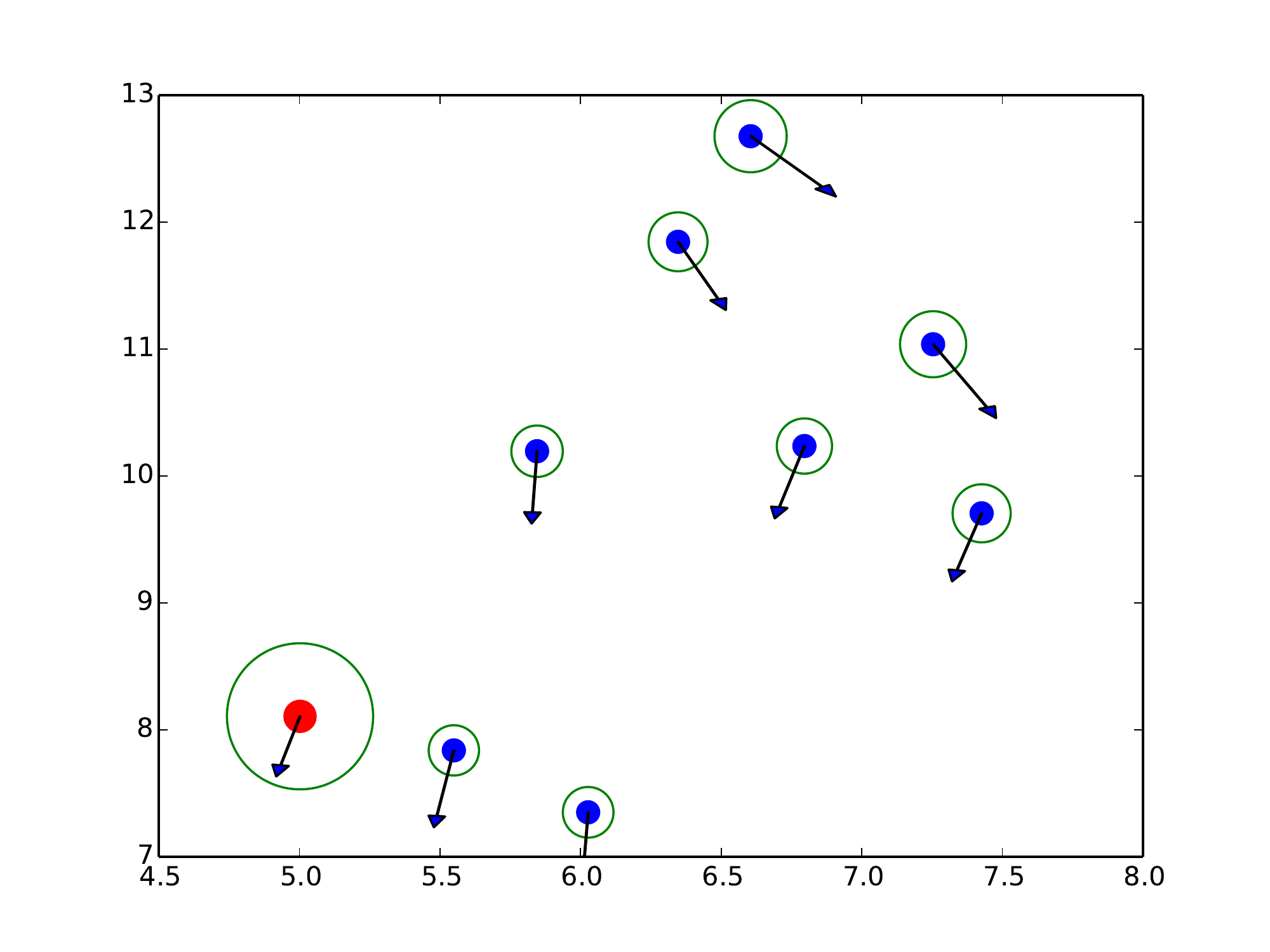}&%
        \includegraphics[width=0.22\linewidth]{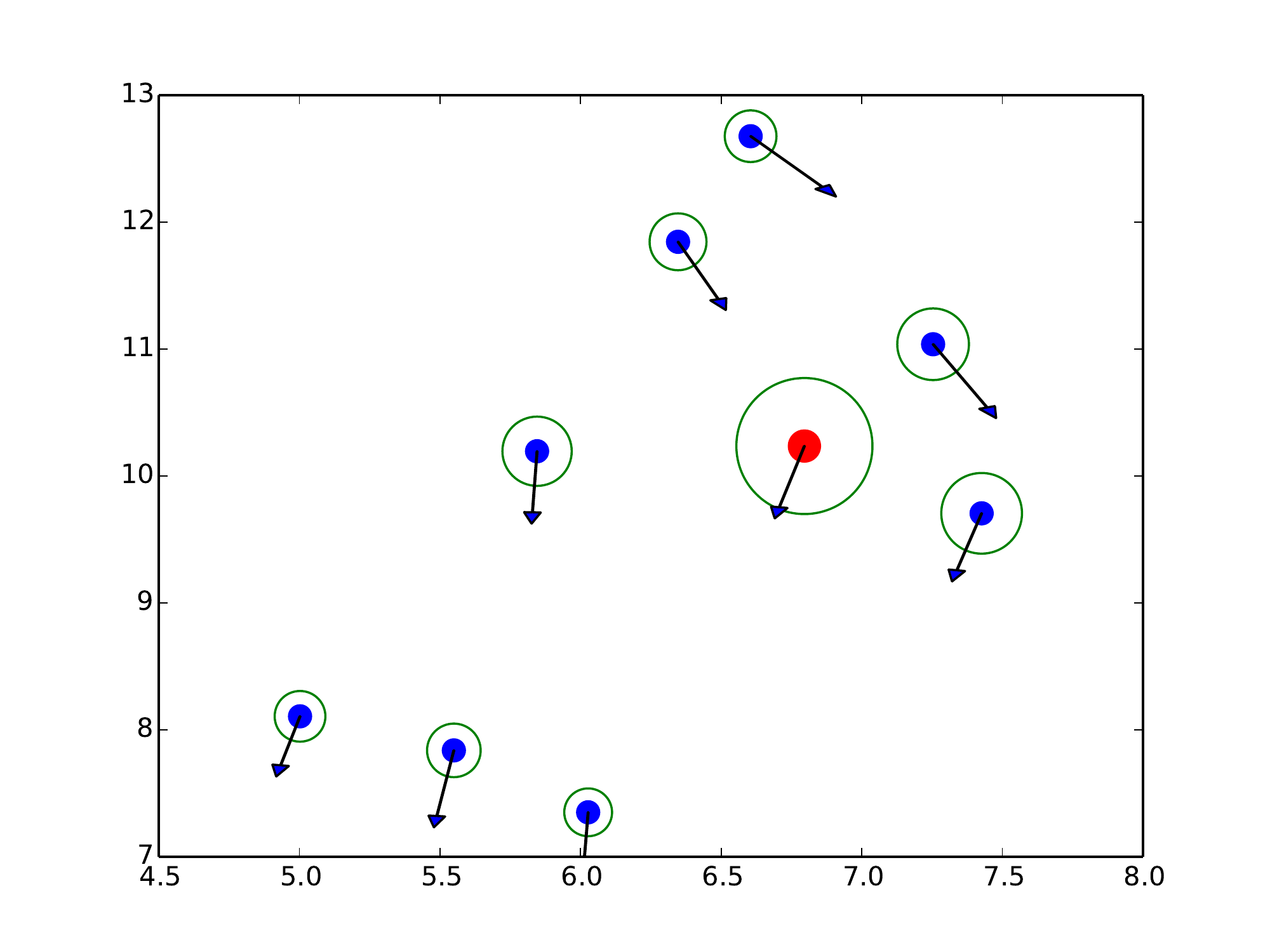}&%
        \includegraphics[width=0.22\linewidth]{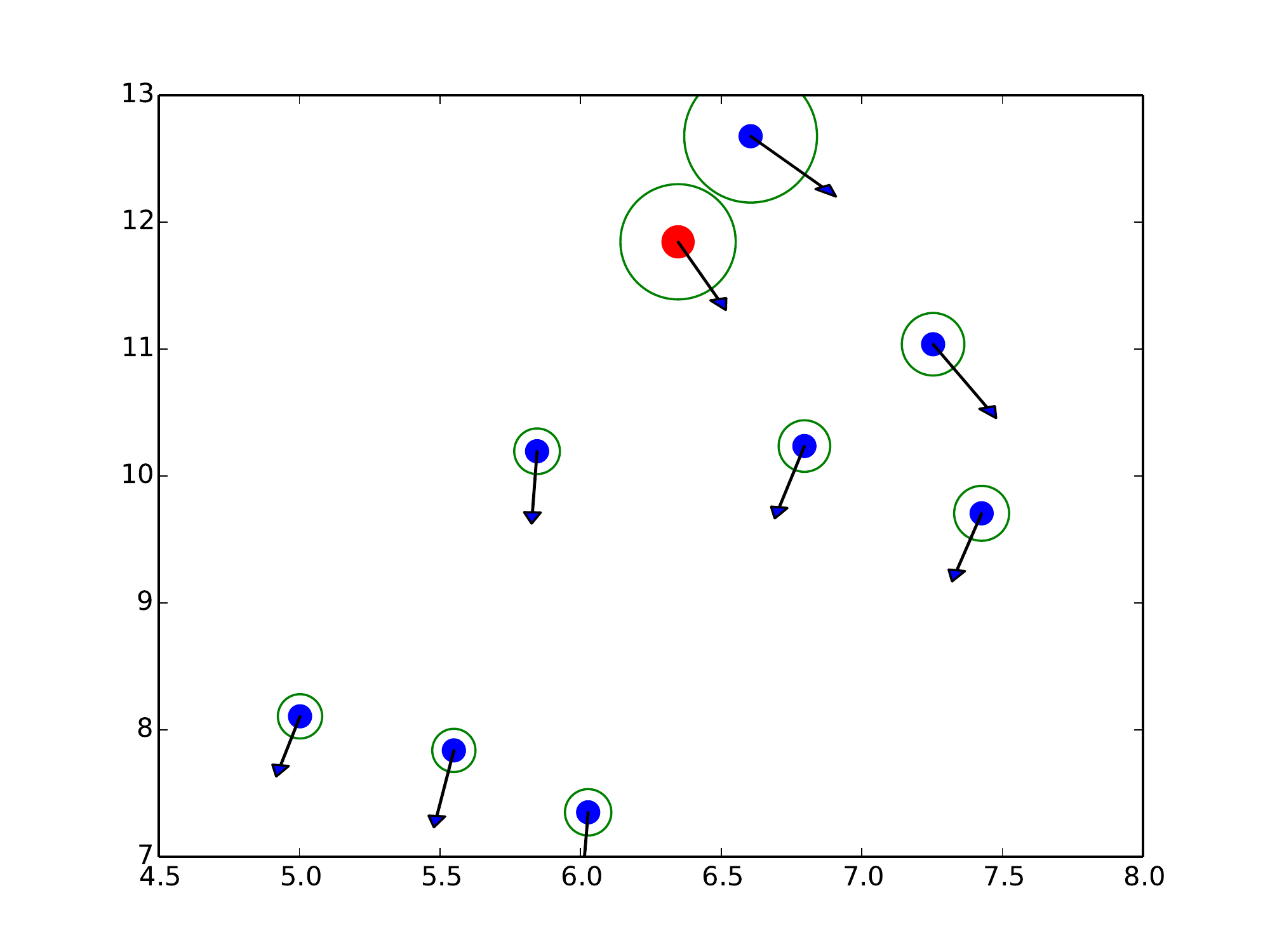}&%
        \includegraphics[width=0.22\linewidth]{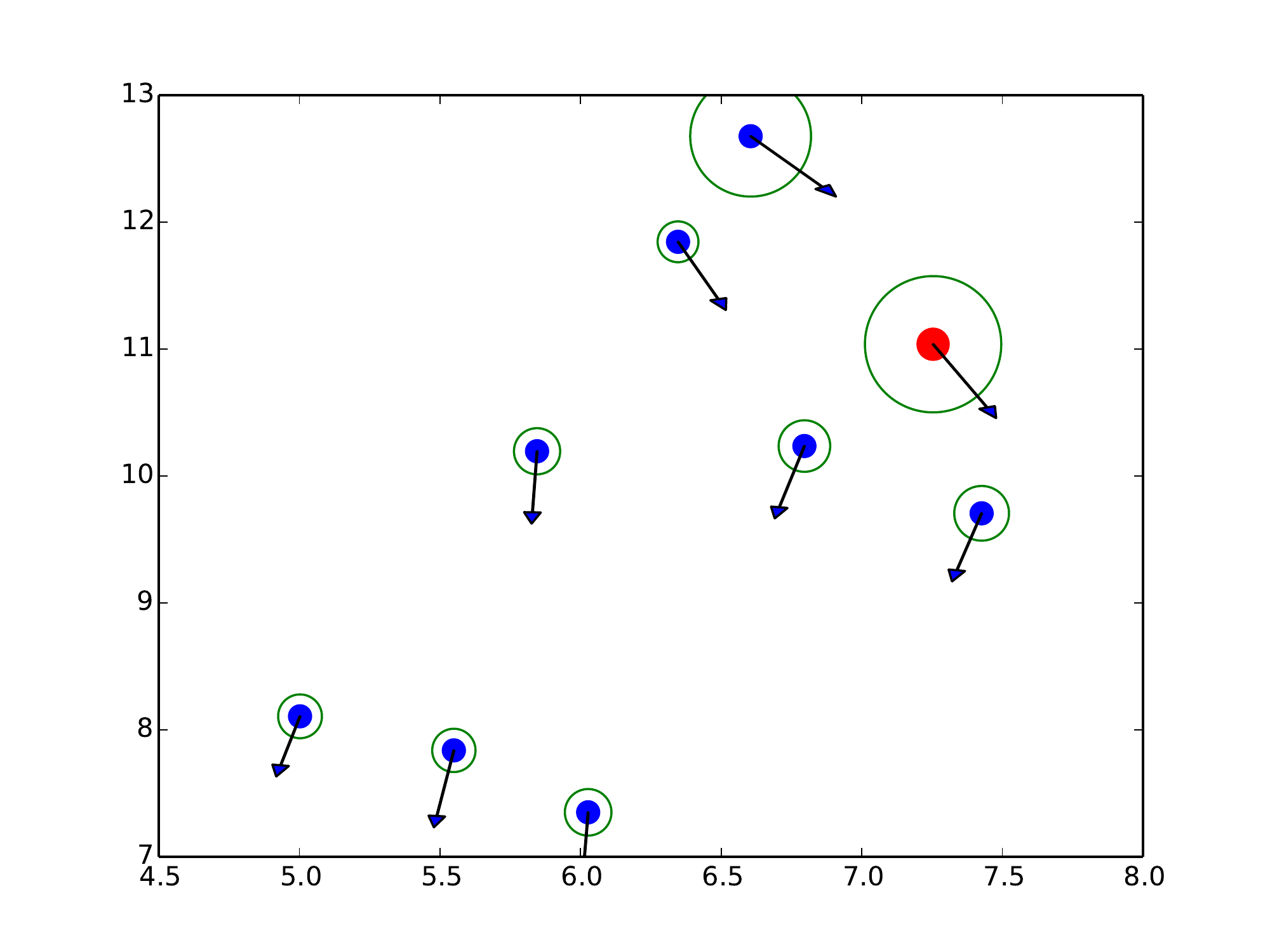}%
        \\  (a) & (b) & (c) &(d)
    \end{tabular}
    \centering
    \caption{Attention visualization of the spatial Transformer in encoder 2. We visualize the attention of all pedestrians with respect to the red dotted pedestrian. The size of circles represents the attention value and bigger circles indicate higher attention. STAR learns reasonable spatial attention, the pedestrians have higher attentions over themselves and their neighbors. }
    \label{fig:att}
\end{figure*}
\begin{itemize}
    \item \emph{STAR is able to predict temporally consistent trajectories}. In Fig.~\ref{fig:dot_vis}.(a), STAR successfully captures the intention and velocity of the single pedestrian, where no social interaction exists.
    \item \emph{STAR successfully extracts the social interaction of the crowd.} We visualize the attention values of the second spatial Transformer in Fig.~\ref{fig:att}. We notice that pedestrians are paying high attention to themselves and the neighbors who might potentially collide with them, e.g., Fig.~\ref{fig:att}.(c) and (d); less attention is paid to spatially far away pedestrians and pedestrians without conflict of intentions, e.g., Fig.~\ref{fig:att}.(a) and (b).
    \item \emph{STAR is able to capture spatio-temporal interaction of the crowd}. In Fig.~\ref{fig:dot_vis}.(b), we can see that the prediction of pedestrian considers the future motions of their neighbors. In addition, STAR better balances the spatial modeling and temporal modeling, compared to SR-LSTM. SR-LSTM potentially overfits on the spatial modeling and often tends to predict curves even when pedestrians are walking straight. This also corresponds to our findings in the quantitative analyses section, that deep predictors overfits onto complex datasets. STAR better alleviates this issue with the spatial-temporal Transformer structure.
    \item \emph{Auxiliary information is required for more accurate trajectory prediction.} Although STAR achieves the SOTA results, prediction can be still inaccurate occasionally, e.g., Fig.~\ref{fig:dot_vis}.(d). The pedestrian takes a sharp turn, which makes it impossible to predict future trajectory purely based on the history of locations. For future work, additional information, e.g., environment setup or map, should be used to provide extra information for prediction.
\end{itemize}

\subsection{Ablation Studies}
We conduct extensive ablation studies on all 5 datasets to understand the influence of each STAR component. Specifically, we choose deterministic STAR to remove the influence of random sample and focus on the effect of the proposed components. The results are presented in Table~\ref{table:ablation}.
\begin{table*}[t!]
\begin{center}
\begin{tabular}{cccc|c|c|c|c|c|c}
\hline %
\multicolumn{4}{c|}{Components} & \multicolumn{6}{c}{Performance (ADE/FDE)} \\
\hline
&SP & TP & GM & ETH  & HOTEL & ZARA1 & ZARA2 & UNIV &AVG\\
\hline
\hline
(1)& GCN & STAR & \checkmark &3.06/5.57 &0.99/1.80 & 2.49/4.58 &1.37/2.52 &1.38/2.47 & 1.86 /3.34\\
(2)& GAT & STAR & \checkmark &0.64/1.25 &0.34/0.72 & 0.47/1.09 &0.37/0.86 &0.55/1.19 &0.48/1.02\\
(3)& MHA & STAR & \checkmark &0.58/1.15 &\textbf{0.25}/\textbf{0.48} & 0.50/0.98 & 0.35/0.76 &0.60/1.24 &0.56/0.92\\
(4)& STAR & LSTM & - &0.66/1.29 &0.34/0.68 &0.45/0.96 &0.34/0.74 &0.60/1.29 &0.48/0.99\\
(5)& STAR & STAR &$\times$ & 0.60/1.18 &0.28/0.60 & 0.53/1.13 & 0.36/0.76 & 0.57/1.20 &0.47/0.97\\
(6) & VSTAR & VSTAR &\checkmark&0.61/1.18 & 0.29/0.56 &0.48/1.00 &0.36/0.76 &0.58/1.24 &0.46/0.95\\
(7)& STAR & STAR &\checkmark&\textbf{0.56}/\textbf{1.11} & 0.26/0.50 & \textbf{0.41}/\textbf{0.90} &\textbf{0.31}/\textbf{0.71} & \textbf{0.52}/\textbf{1.15} &\textbf{0.41}/\textbf{0.87}\\
\hline
\end{tabular}
\end{center}
\caption{Ablation Study on SR-LSTM. We replace components in STAR with existing works. \textbf{SP} denotes spaital encoder. \textbf{TP} denotes temporal encoder. \textbf{GM} denotes Graph Memory. \textbf{GAT} denotes Graph Attention Network\cite{velivckovic2017graph}, \textbf{MHA} denotes Multi-Head Additive attention\cite{chen2019PAT}.\textbf{STAR} denotes components in original STAR. \textbf{VSTAR} denoets simplified STAR without encoder2. }
\label{table:ablation}
\end{table*}
\begin{itemize}
    \item \emph{The temporal Transformer improves the temporal modeling of pedestrian dynamics compared to RNNs.} In (4) and (5), we remove the graph memory and fix the STAR for spatial encoding. The temporal prediction ability of these two models is only dependent on their temporal encoders, LSTM for (4) and STAR for (5). We observe that the model with temporal Transformer encoding outperforms LSTM in its overall performance, which suggests that Transformers provide a better temporal modeling ability compared to RNNs.
    \item \emph{TGConv outperforms the other graph convolution methods on crowd motion modeling.} In (1), (2), (3) and (7), we change the spatial encoders and compare the spatial Transformer by TGConv (7) with the GCN~\cite{kipf2016semi}, GATConv~\cite{velivckovic2017graph} and the multi-head additive graph convolution~\cite{chen2019PAT}. We observe that TGConv, under the scenario of crowd modeling, achieves higher performance gain compared to the other two alternative attention-based graph convolutions.
    \item \emph{Interleaving spatial and temporal Transformer is able to better extract spatio-temporal correlations.} In (6) and (7), we observe that the two encoder structures proposed in the STAR framework (7), generally outperforms the single encoder structure (6). This empirical performance gain potentially suggests that interleaving the spatial and temporal Transformers is able to extract more complex spatio-temporal interactions of pedestrians.
    \item \emph{Graph memory gives a smoother temporal embedding and improves performance.} In (5) and (7), we verify the embedding smoothing ability of the graph memory module, where (5) is the STAR variant without GM. We first noticed that graph memory improves the performance of STAR on all datasets. In addition, we noticed that on ZARA1, where the spatial interaction is simple and temporal consistency prediction is more important, graph memory improves (6) to (7) by the largest margin. According to the empirical evidence, we can conclude that the embedding smoothing of graph memory is able to improve the overall temporal modeling for STAR.
\end{itemize}

\section{Conclusion}
We have introduced STAR, a framework for spatio-temporal crowd trajectory prediction with only attention mechanisms. 
STAR consists of two encoder modules, composed of spatial Transformers and temporal Transformers. We also have introduced TGConv, a novel powerful Transformer based graph convolution mechanism. STAR, using only attention mechanisms, achieves SOTA performance on 5 commonly used datasets.

STAR makes prediction only with the past trajectories, which might fail to detect the unpredictable sharp turns. Additional information, e.g., environment configuration, could be incorporated into the framework to solve this issue.

STAR framework and TGConv are not limited to trajectory prediction. They can be applied to any graph learning task. We leave it for future study.

\clearpage
\bibliographystyle{splncs04}

\begin{thebibliography}{10}
\providecommand{\url}[1]{\texttt{#1}}
\providecommand{\urlprefix}{URL }
\providecommand{\doi}[1]{https://doi.org/#1}

\bibitem{alahi2016social}
Alahi, A., Goel, K., Ramanathan, V., Robicquet, A., Fei-Fei, L., Savarese, S.:
  Social lstm: Human trajectory prediction in crowded spaces. In: CVPR (2016)

\bibitem{ba2016layer}
Ba, J.L., Kiros, J.R., Hinton, G.E.: Layer normalization. arXiv preprint
  arXiv:1607.06450  (2016)

\bibitem{bahdanau2014neural}
Bahdanau, D., Cho, K., Bengio, Y.: Neural machine translation by jointly
  learning to align and translate. arXiv preprint arXiv:1409.0473  (2014)

\bibitem{battaglia2016interaction}
Battaglia, P., Pascanu, R., Lai, M., Rezende, D.J., et~al.: Interaction
  networks for learning about objects, relations and physics. In: Advances in
  neural information processing systems (2016)

\bibitem{chen2019PAT}
Chen, B., Barzilay, R., Jaakkola, T.: Path-augmented graph transformer network
  (2019). \doi{10.26434/chemrxiv.8214422}

\bibitem{cho2014learning}
Cho, K., van Merrienboer, B., Gulcehre, C., Bahdanau, D., Bougares, F.,
  Schwenk, H., Bengio, Y.: Learning phrase representations using {RNN}
  encoder--decoder for statistical machine translation. In: Proceedings of the
  2014 Conference on Empirical Methods in Natural Language Processing (2014)

\bibitem{chung2014empirical}
Chung, J., Gulcehre, C., Cho, K., Bengio, Y.: Empirical evaluation of gated
  recurrent neural networks on sequence modeling. arXiv preprint
  arXiv:1412.3555  (2014)

\bibitem{cui2019traffic}
Cui, Z., Henrickson, K., Ke, R., Wang, Y.: Traffic graph convolutional
  recurrent neural network: A deep learning framework for network-scale traffic
  learning and forecasting. IEEE Transactions on Intelligent Transportation
  Systems  (2019)

\bibitem{defferrard2016convolutional}
Defferrard, M., Bresson, X., Vandergheynst, P.: Convolutional neural networks
  on graphs with fast localized spectral filtering. In: Advances in neural
  information processing systems (2016)

\bibitem{devlin2018bert}
Devlin, J., Chang, M.W., Lee, K., Toutanova, K.: Bert: Pre-training of deep
  bidirectional transformers for language understanding. arXiv preprint
  arXiv:1810.04805  (2018)

\bibitem{fan2019graph}
Fan, W., Ma, Y., Li, Q., He, Y., Zhao, E., Tang, J., Yin, D.: Graph neural
  networks for social recommendation. In: WWW (2019)

\bibitem{fang2019scene}
Fang, K., Toshev, A., Fei-Fei, L., Savarese, S.: Scene memory transformer for
  embodied agents in long-horizon tasks. In: CVPR (2019)

\bibitem{ferrer2013robot}
Ferrer, G., Garrell, A., Sanfeliu, A.: Robot companion: A social-force based
  approach with human awareness-navigation in crowded environments. In: IROS
  (2013)

\bibitem{forster2007rnn}
F{\"o}rster, A., Graves, A., Schmidhuber, J.: Rnn-based learning of compact
  maps for efficient robot localization. In: ESANN (2007)

\bibitem{gilmer2017neural}
Gilmer, J., Schoenholz, S.S., Riley, P.F., Vinyals, O., Dahl, G.E.: Neural
  message passing for quantum chemistry. In: ICML (2017)

\bibitem{gupta2018social}
Gupta, A., Johnson, J., Fei-Fei, L., Savarese, S., Alahi, A.: Social gan:
  Socially acceptable trajectories with generative adversarial networks. In:
  CVPR (2018)

\bibitem{hajiramezanali2019variational}
Hajiramezanali, E., Hasanzadeh, A., Narayanan, K., Duffield, N., Zhou, M.,
  Qian, X.: Variational graph recurrent neural networks. In: Advances in Neural
  Information Processing Systems (2019)

\bibitem{helbing2005self}
Helbing, D., Buzna, L., Johansson, A., Werner, T.: Self-organized pedestrian
  crowd dynamics: Experiments, simulations, and design solutions.
  Transportation science  (2005)

\bibitem{helbing1995social}
Helbing, D., Molnar, P.: Social force model for pedestrian dynamics. Physical
  review E  (1995)

\bibitem{hochreiter1997long}
Hochreiter, S., Schmidhuber, J.: Long short-term memory. Neural computation
  (1997)

\bibitem{huang2019stgat}
Huang, Y., Bi, H., Li, Z., Mao, T., Wang, Z.: Stgat: Modeling spatial-temporal
  interactions for human trajectory prediction. In: ICCV (2019)

\bibitem{ivanovic2019trajectron}
Ivanovic, B., Pavone, M.: The trajectron: Probabilistic multi-agent trajectory
  modeling with dynamic spatiotemporal graphs. In: ICCV (2019)

\bibitem{karkus2019differentiable}
Karkus, P., Ma, X., Hsu, D., Kaelbling, L.P., Lee, W.S., Lozano-P{\'e}rez, T.:
  Differentiable algorithm networks for composable robot learning. arXiv
  preprint arXiv:1905.11602  (2019)

\bibitem{kipf2016semi}
Kipf, T.N., Welling, M.: Semi-supervised classification with graph
  convolutional networks. arXiv preprint arXiv:1609.02907  (2016)

\bibitem{kuderer2012feature}
Kuderer, M., Kretzschmar, H., Sprunk, C., Burgard, W.: Feature-based prediction
  of trajectories for socially compliant navigation. In: RSS (2012)

\bibitem{lan2019albert}
Lan, Z., Chen, M., Goodman, S., Gimpel, K., Sharma, P., Soricut, R.: Albert: A
  lite bert for self-supervised learning of language representations. arXiv
  preprint arXiv:1909.11942  (2019)

\bibitem{li2015gated}
Li, Y., Tarlow, D., Brockschmidt, M., Zemel, R.: Gated graph sequence neural
  networks. arXiv preprint arXiv:1511.05493  (2015)

\bibitem{li2018learning}
Li, Y., Wu, J., Tedrake, R., Tenenbaum, J.B., Torralba, A.: Learning particle
  dynamics for manipulating rigid bodies, deformable objects, and fluids. arXiv
  preprint arXiv:1810.01566  (2018)

\bibitem{lim2019temporal}
Lim, B., Arik, S.O., Loeff, N., Pfister, T.: Temporal fusion transformers for
  interpretable multi-horizon time series forecasting. arXiv preprint
  arXiv:1912.09363  (2019)

\bibitem{liu2019transformer}
Liu, J., Lin, H., Liu, X., Xu, B., Ren, Y., Diao, Y., Yang, L.:
  Transformer-based capsule network for stock movement prediction. In:
  Proceedings of the First Workshop on Financial Technology and Natural
  Language Processing (2019)

\bibitem{liu2019chemi}
Liu, K., Sun, X., Jia, L., Ma, J., Xing, H., Wu, J., Gao, H., Sun, Y.,
  Boulnois, F., Fan, J.: Chemi-net: a molecular graph convolutional network for
  accurate drug property prediction. International journal of molecular
  sciences  (2019)

\bibitem{lohner2010modeling}
L{\"o}hner, R.: On the modeling of pedestrian motion. Applied Mathematical
  Modelling  (2010)

\bibitem{luo2019gamma}
Luo, Y., Cai, P.: Gamma: A general agent motion prediction model for autonomous
  driving. arXiv preprint arXiv:1906.01566  (2019)

\bibitem{luo2018porca}
Luo, Y., Cai, P., Bera, A., Hsu, D., Lee, W.S., Manocha, D.: Porca: Modeling
  and planning for autonomous driving among many pedestrians. IEEE Robotics and
  Automation Letters  (2018)

\bibitem{ma2017beep}
Ma, X., Gao, X., Chen, G.: Beep: A bayesian perspective early stage event
  prediction model for online social networks. In: ICDM (2017)

\bibitem{ma2019particle}
Ma, X., Karkus, P., Hsu, D., Lee, W.S.: Particle filter recurrent neural
  networks. arXiv preprint arXiv:1905.12885  (2019)

\bibitem{ma2020discriminative}
Ma, X., Karkus, P., Hsu, D., Lee, W.S., Ye, N.: Discriminative particle filter
  reinforcement learning for complex partial observations. arXiv preprint
  arXiv:2002.09884  (2020)

\bibitem{mayuexin2019trafficpredict}
Ma, Y., Zhu, X., Zhang, S., Yang, R., Wang, W., Manocha, D.: Trafficpredict:
  Trajectory prediction for heterogeneous traffic-agents. AAAI  (2019)

\bibitem{miao2015eesen}
Miao, Y., Gowayyed, M., Metze, F.: Eesen: End-to-end speech recognition using
  deep rnn models and wfst-based decoding. In: ASRU (2015)

\bibitem{Sadeghian2019sophie}
Sadeghian, A., Kosaraju, V., Sadeghian, A., Hirose, N., Rezatofighi, H.,
  Savarese, S.: Sophie: An attentive gan for predicting paths compliant to
  social and physical constraints. In: CVPR (2019)

\bibitem{sutskever2014sequence}
Sutskever, I., Vinyals, O., Le, Q.V.: Sequence to sequence learning with neural
  networks. In: Advances in neural information processing systems (2014)

\bibitem{van2011reciprocal}
Van Den~Berg, J., Guy, S.J., Lin, M., Manocha, D.: Reciprocal n-body collision
  avoidance. In: Robotics research (2011)

\bibitem{vaswani2017attention}
Vaswani, A., Shazeer, N., Parmar, N., Uszkoreit, J., Jones, L., Gomez, A.N.,
  Kaiser, {\L}., Polosukhin, I.: Attention is all you need. In: Advances in
  neural information processing systems (2017)

\bibitem{velivckovic2017graph}
Veli{\v{c}}kovi{\'c}, P., Cucurull, G., Casanova, A., Romero, A., Lio, P.,
  Bengio, Y.: Graph attention networks. arXiv preprint arXiv:1710.10903  (2017)

\bibitem{vemula2018social}
Vemula, A., Muelling, K., Oh, J.: Social attention: Modeling attention in human
  crowds. In: ICRA (2018)

\bibitem{xiong2018microsoft}
Xiong, W., Wu, L., Alleva, F., Droppo, J., Huang, X., Stolcke, A.: The
  {M}icrosoft 2017 conversational speech recognition system. In: Proceedings of
  the IEEE International Conference on Acoustics, Speech and Signal Processing
  (2018)

\bibitem{xu2018powerful}
Xu, K., Hu, W., Leskovec, J., Jegelka, S.: How powerful are graph neural
  networks? arXiv preprint arXiv:1810.00826  (2018)

\bibitem{xu2018encoding}
Xu, Y., Piao, Z., Gao, S.: Encoding crowd interaction with deep neural network
  for pedestrian trajectory prediction. In: CVPR (2018)

\bibitem{xu2018cidnn}
Xu, Y., Piao, Z., Gao, S.: Encoding crowd interaction with deep neural network
  for pedestrian trajectory prediction. In: CVPR (2018)

\bibitem{yang2019xlnet}
Yang, Z., Dai, Z., Yang, Y., Carbonell, J., Salakhutdinov, R.R., Le, Q.V.:
  Xlnet: Generalized autoregressive pretraining for language understanding. In:
  Advances in neural information processing systems (2019)

\bibitem{yi2016pedestrian}
Yi, S., Li, H., Wang, X.: Pedestrian behavior understanding and prediction with
  deep neural networks. In: ECCV (2016)

\bibitem{young2018recent}
Young, T., Hazarika, D., Poria, S., Cambria, E.: Recent trends in deep learning
  based natural language processing. ieee Computational intelligenCe magazine
  (2018)

\bibitem{zhang2019sr}
Zhang, P., Ouyang, W., Zhang, P., Xue, J., Zheng, N.: Sr-lstm: State refinement
  for lstm towards pedestrian trajectory prediction. In: CVPR (2019)

\end{thebibliography}

\clearpage
\appendix
\section*{Additional Attention Visualization}
\begin{figure*}[!htb]
	\fontsize{6}{11}\selectfont
	\centering
	\begin{tabular}{c c c c}
		\includegraphics[width=0.24\linewidth]{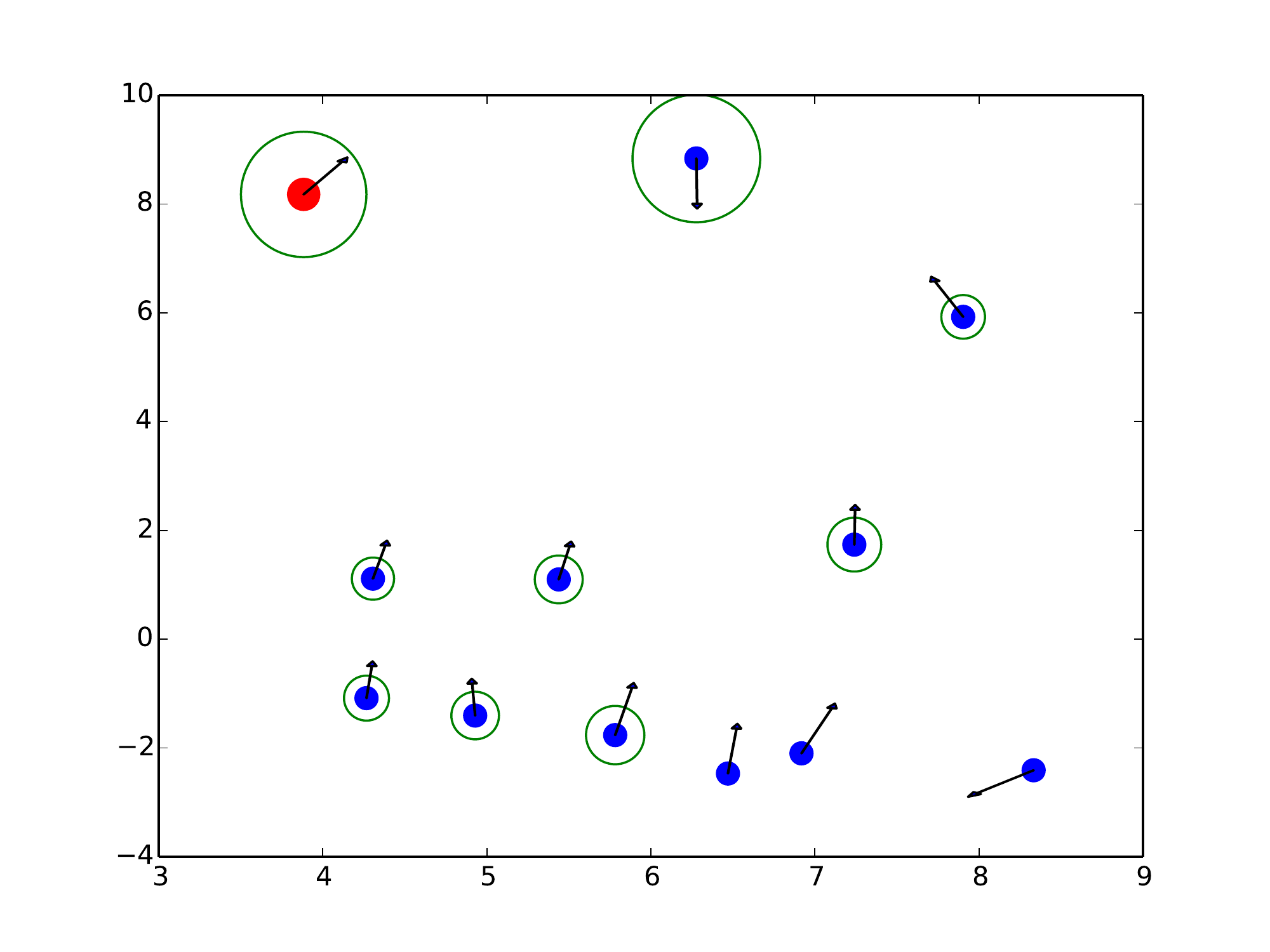}&%
		\includegraphics[width=0.24\linewidth]{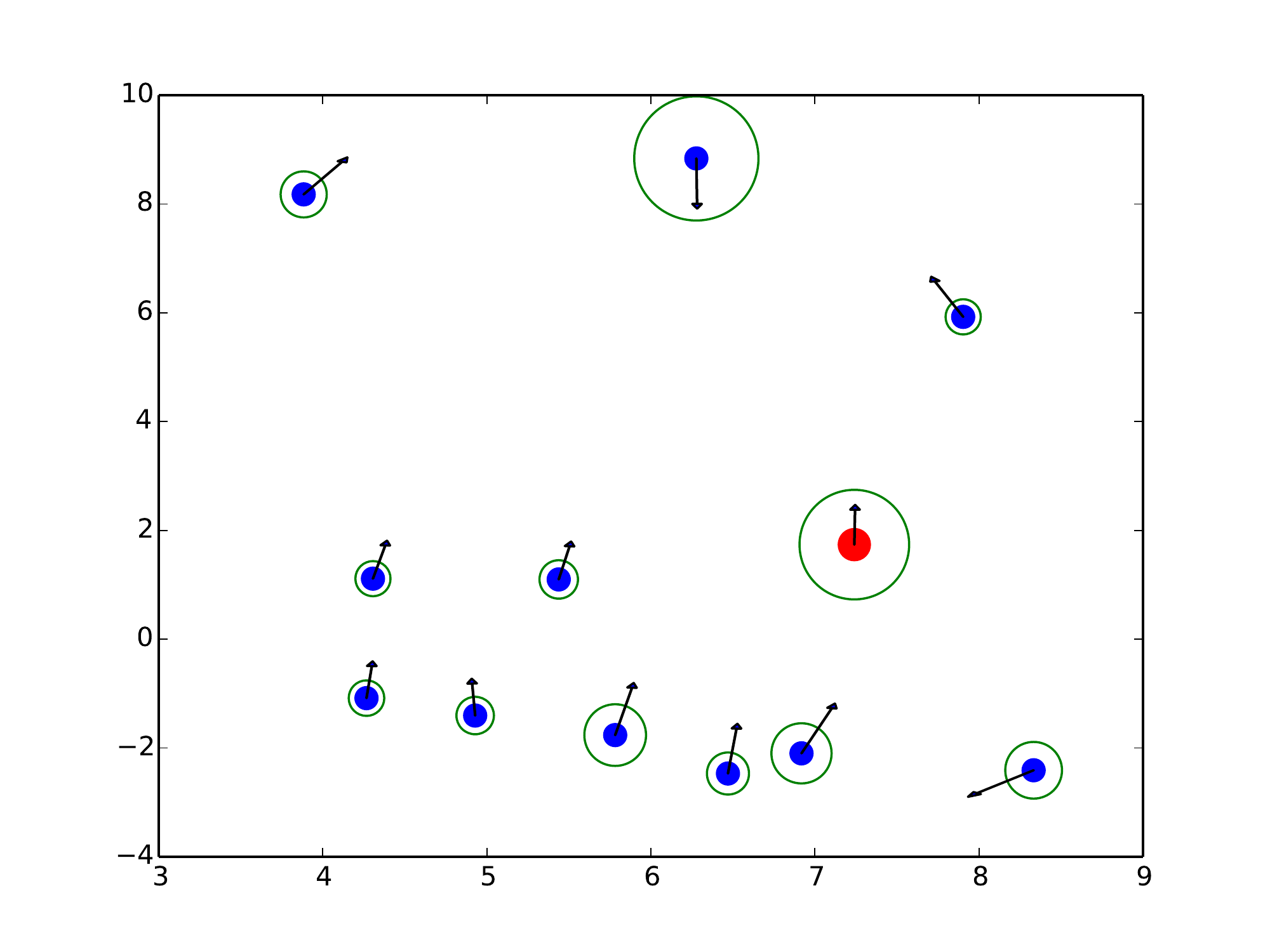}&%
		\includegraphics[width=0.24\linewidth]{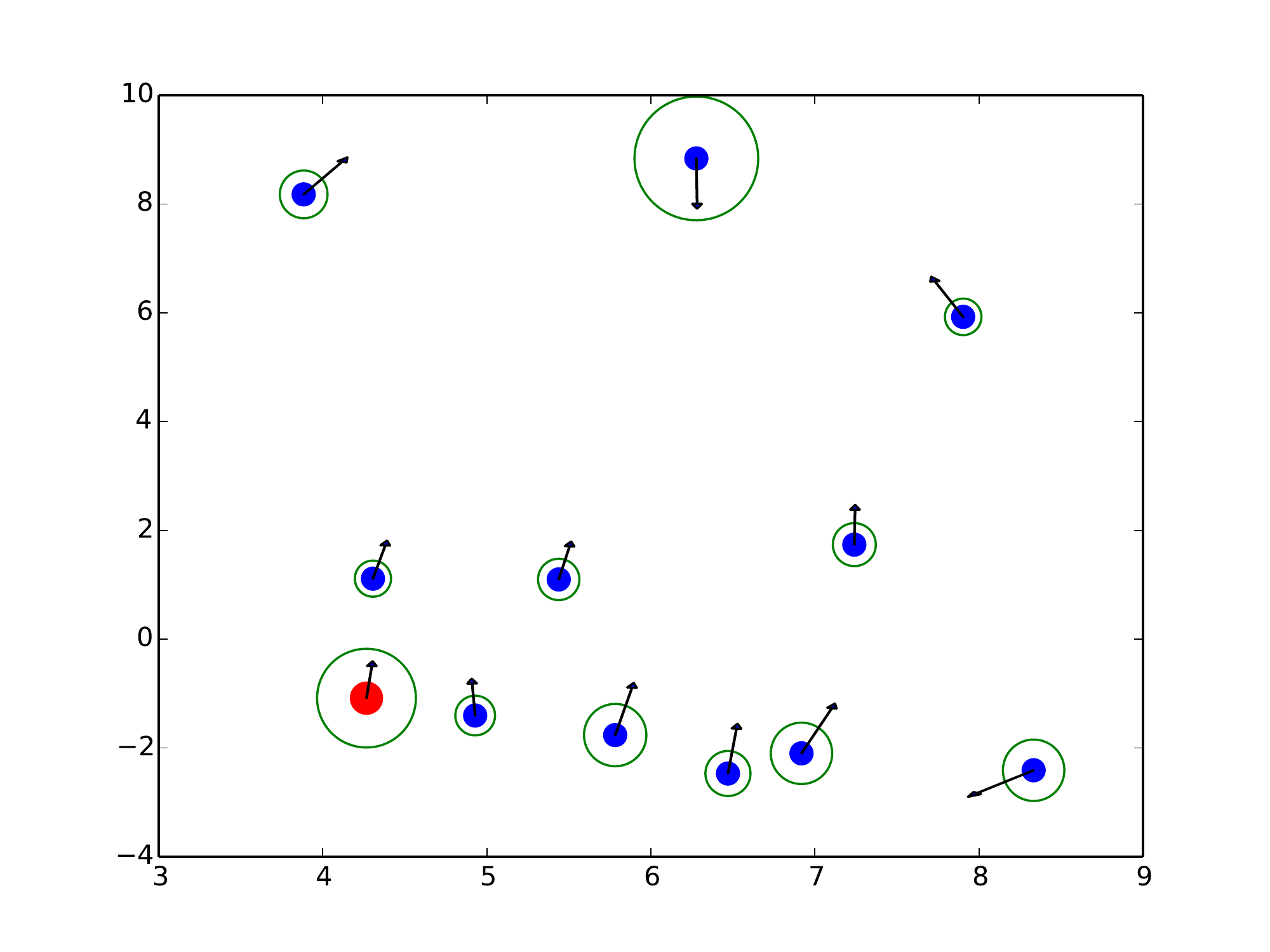}&%
		\includegraphics[width=0.24\linewidth]{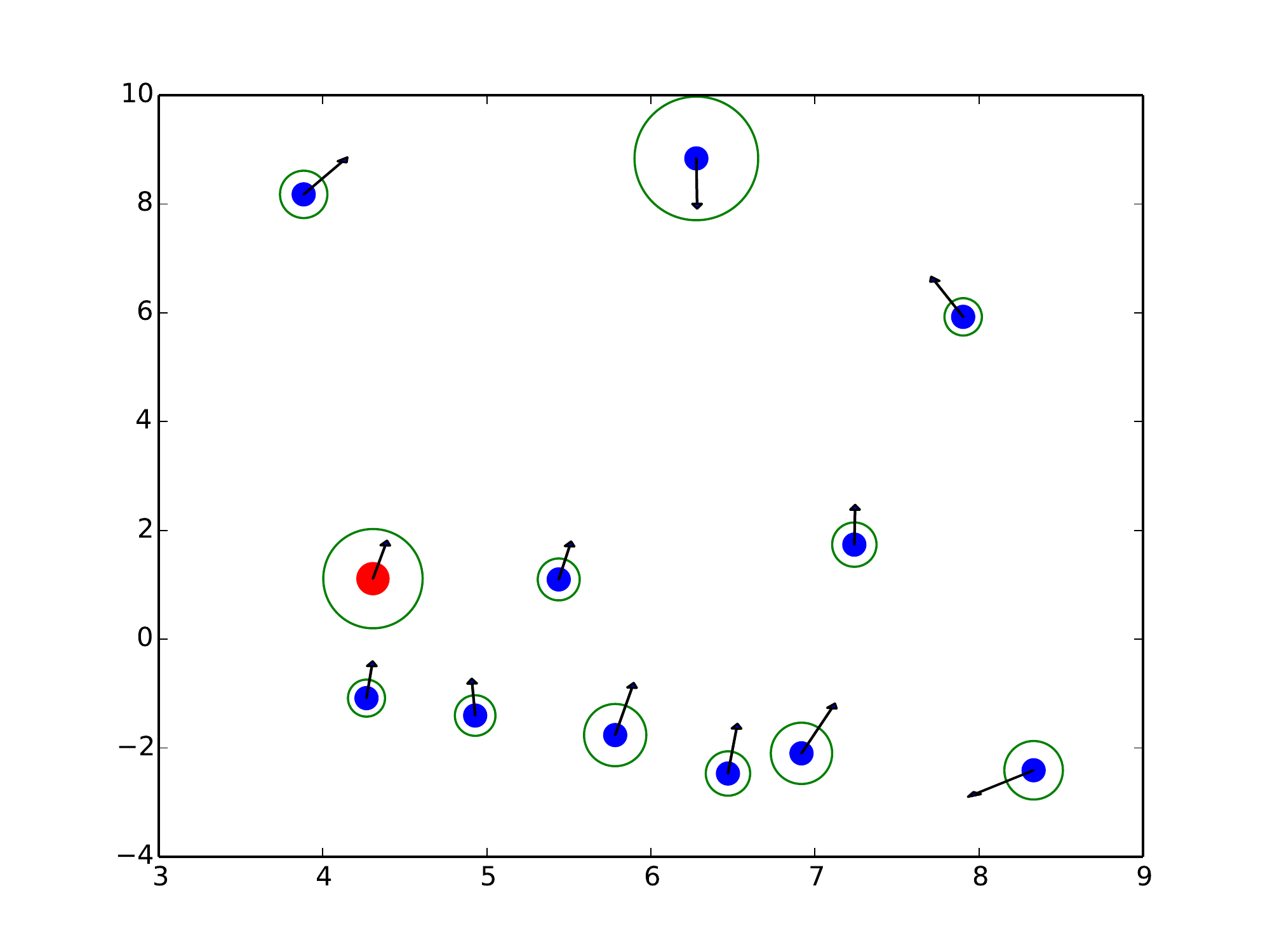}
		\\ \includegraphics[width=0.24\linewidth]{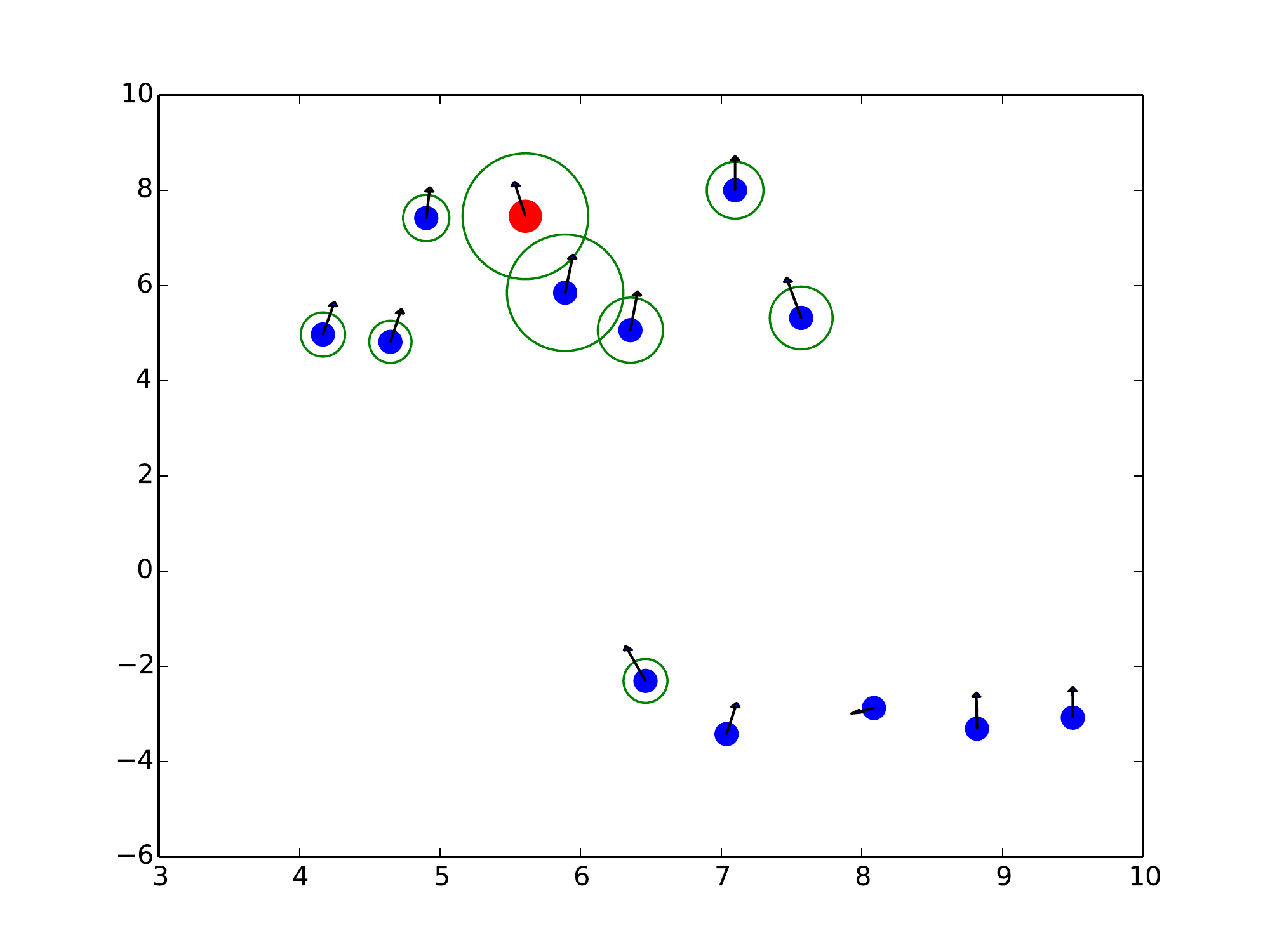}&%
		\includegraphics[width=0.24\linewidth]{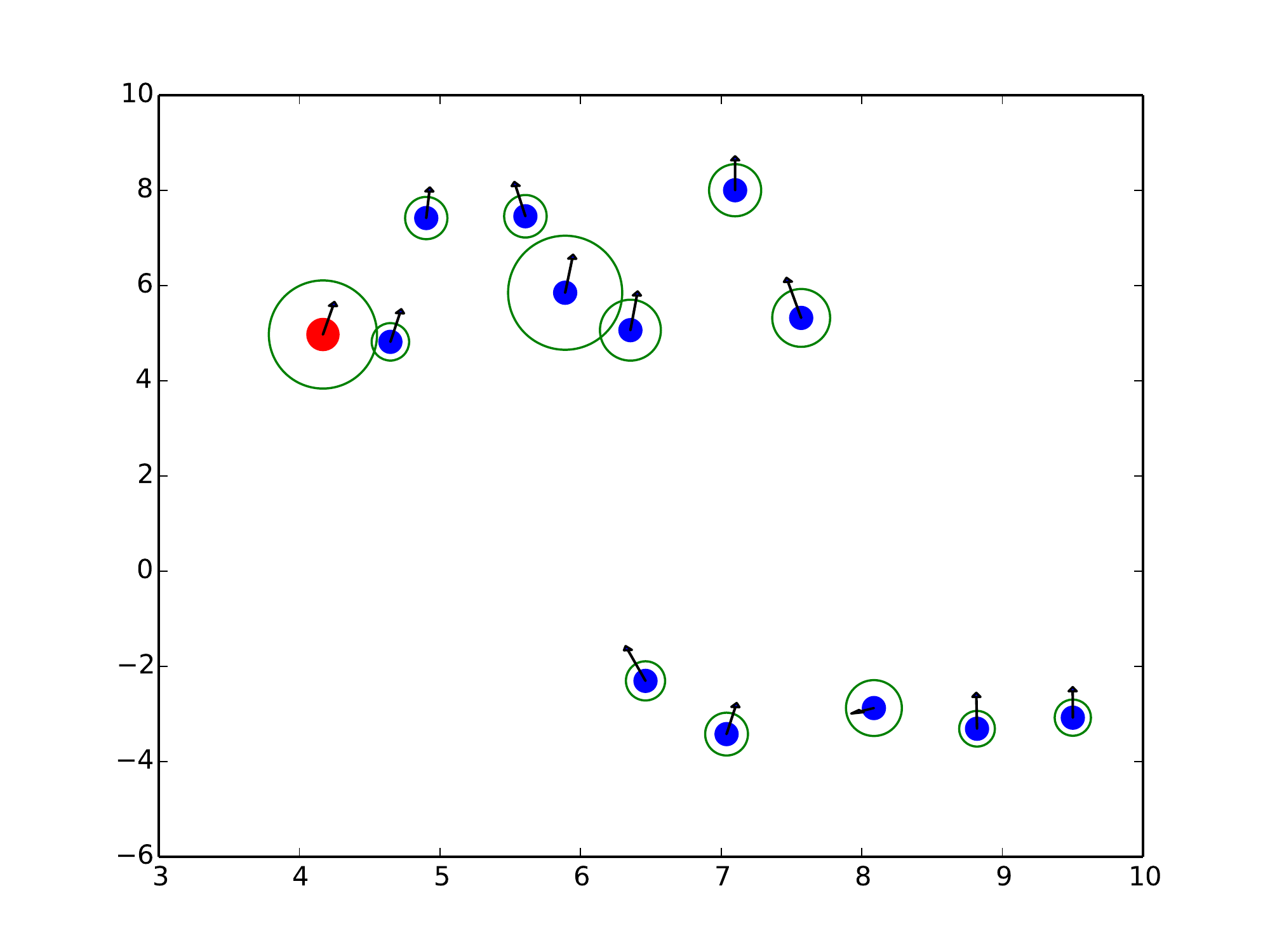}&%
		\includegraphics[width=0.24\linewidth]{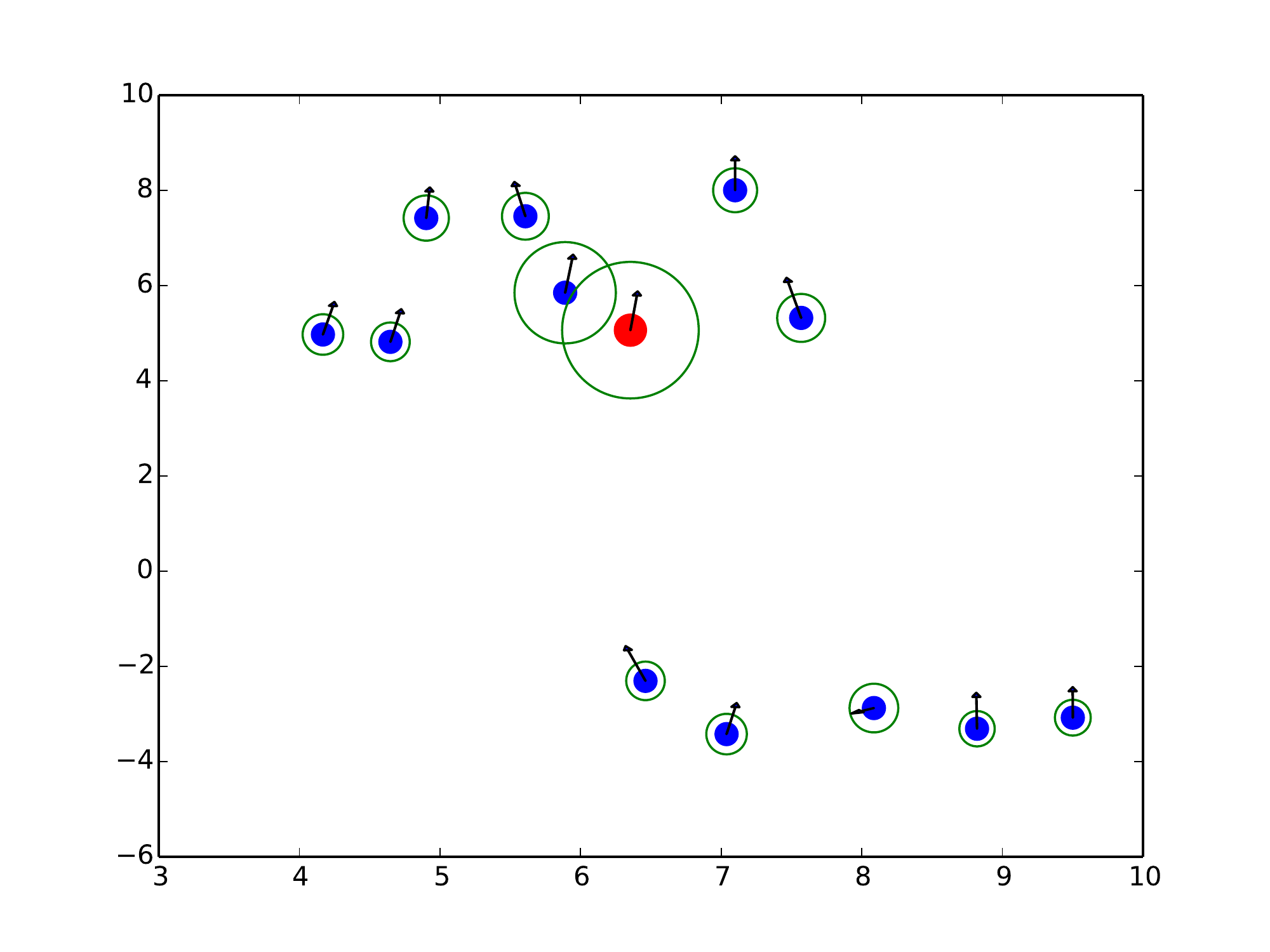}&%
		\includegraphics[width=0.24\linewidth]{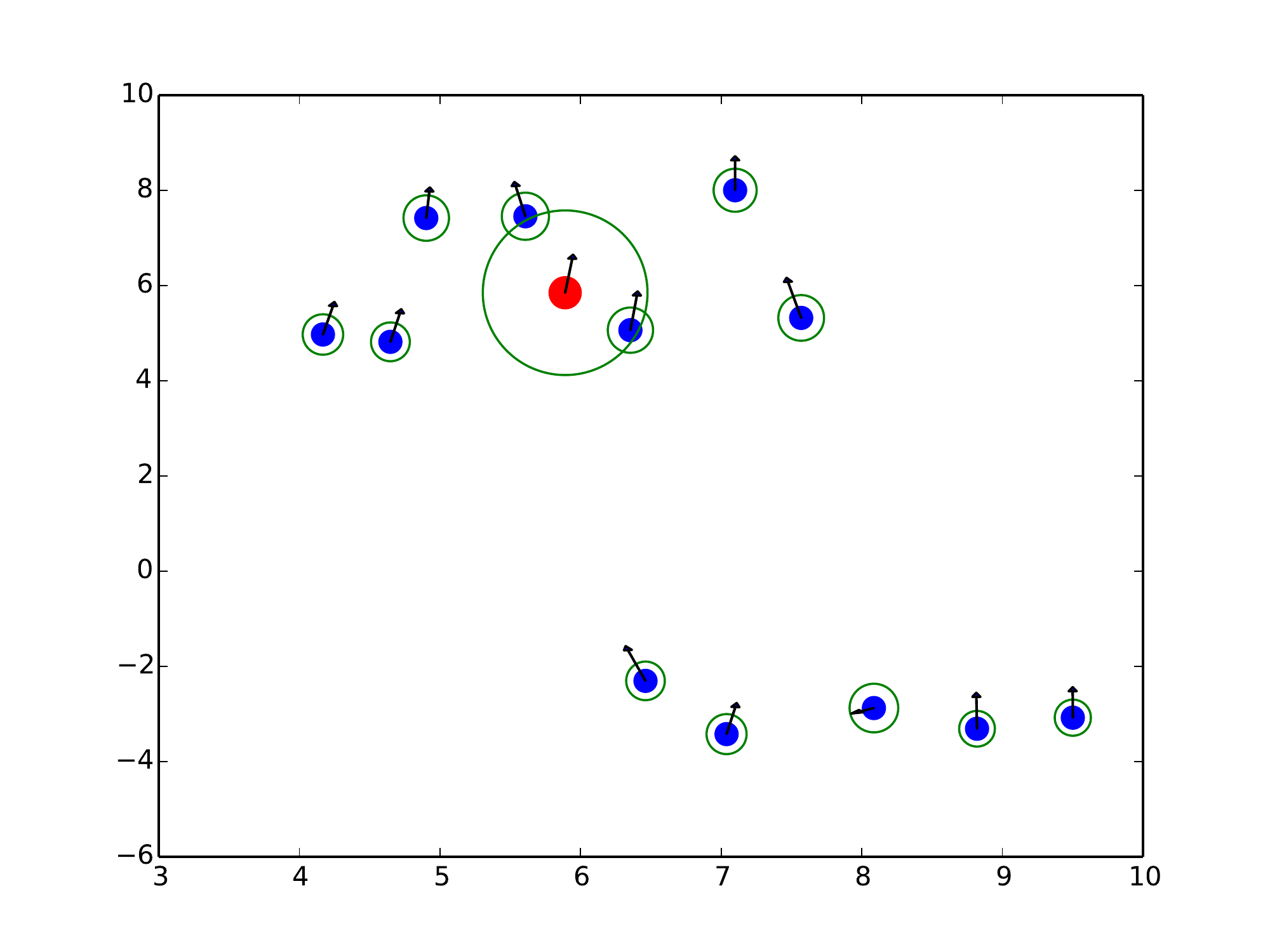}%
		\\ \includegraphics[width=0.24\linewidth]{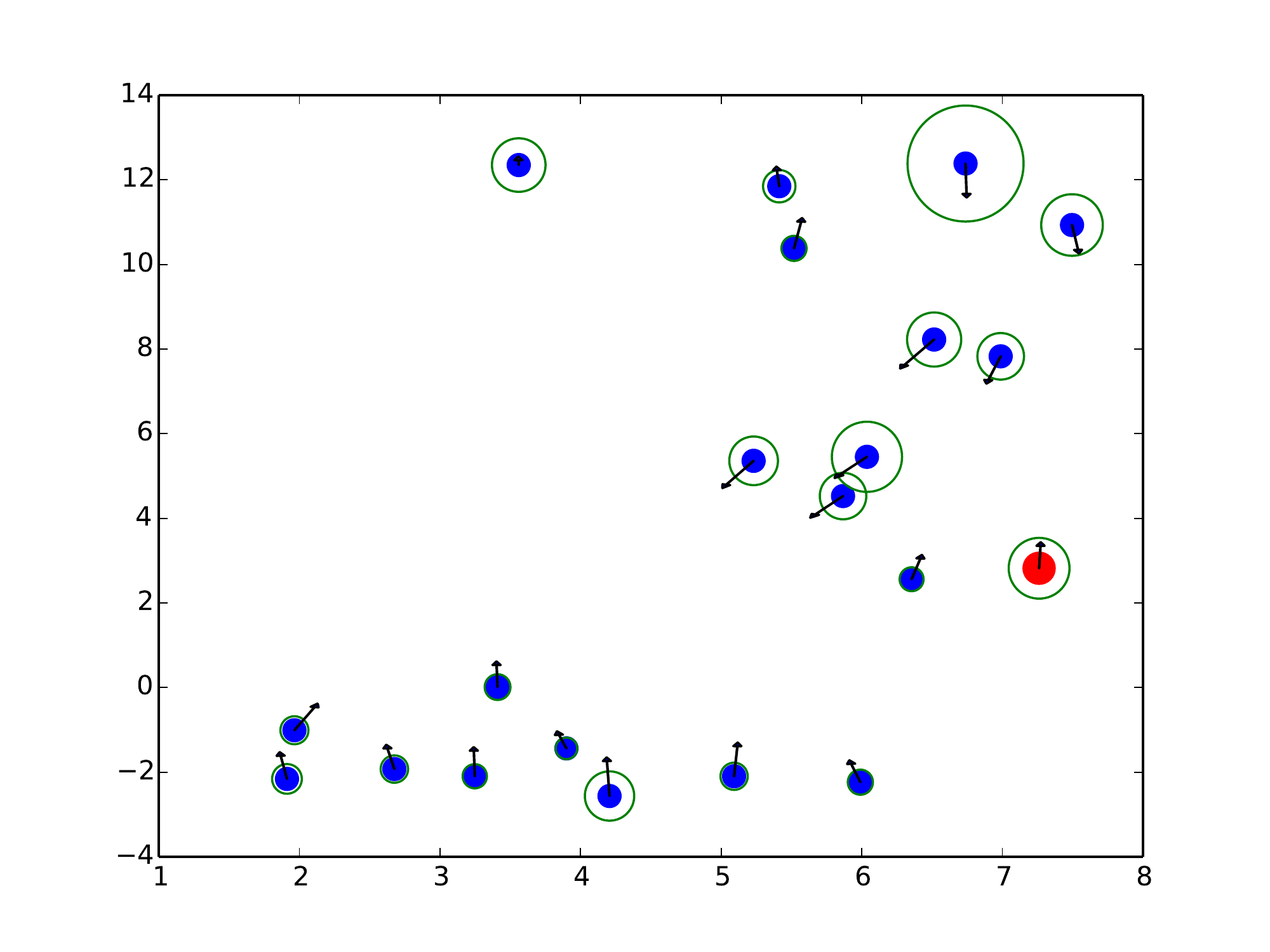}&%
		\includegraphics[width=0.24\linewidth]{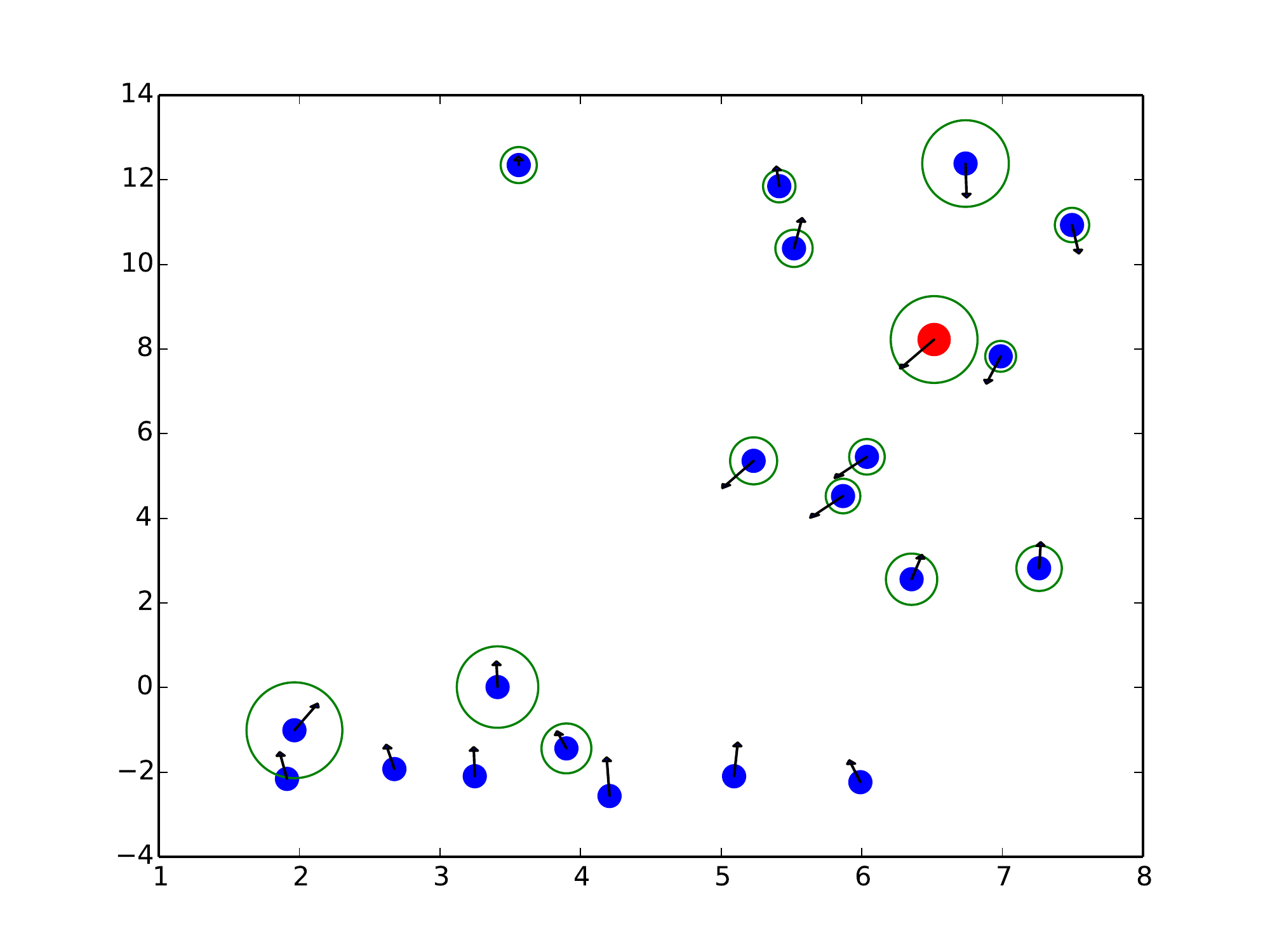}&%
		\includegraphics[width=0.24\linewidth]{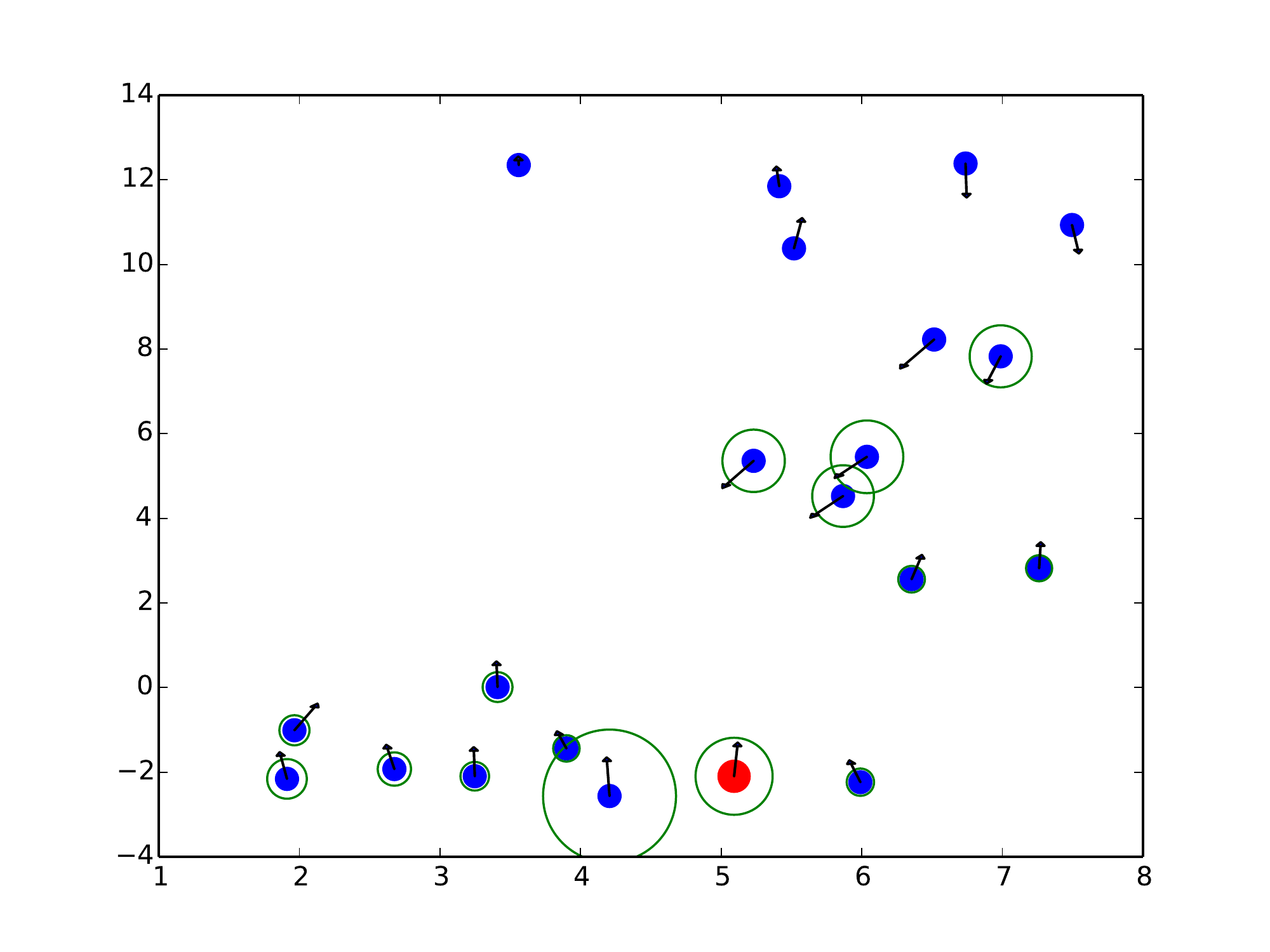}&%
		\includegraphics[width=0.24\linewidth]{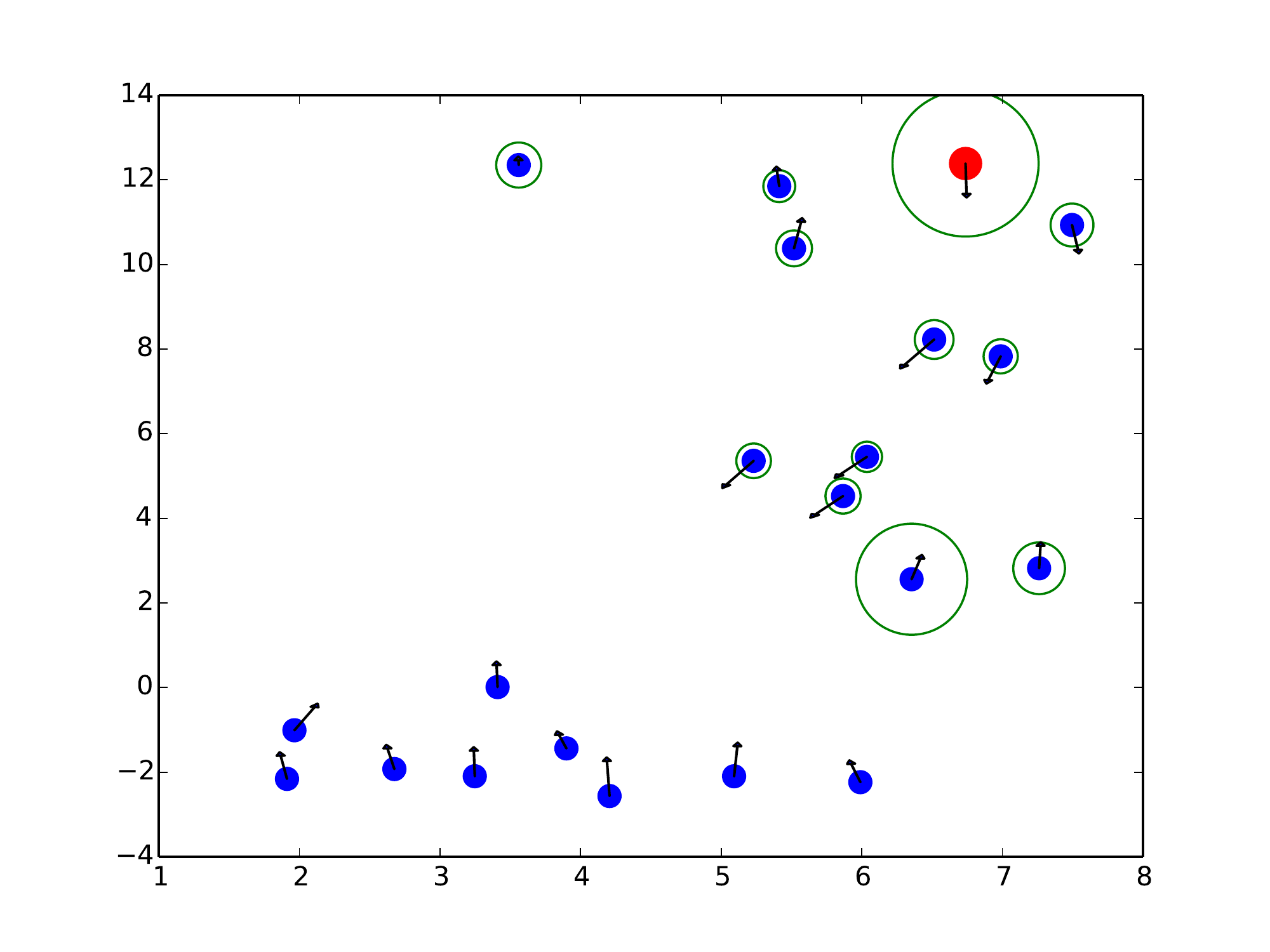}%
		\\ \includegraphics[width=0.24\linewidth]{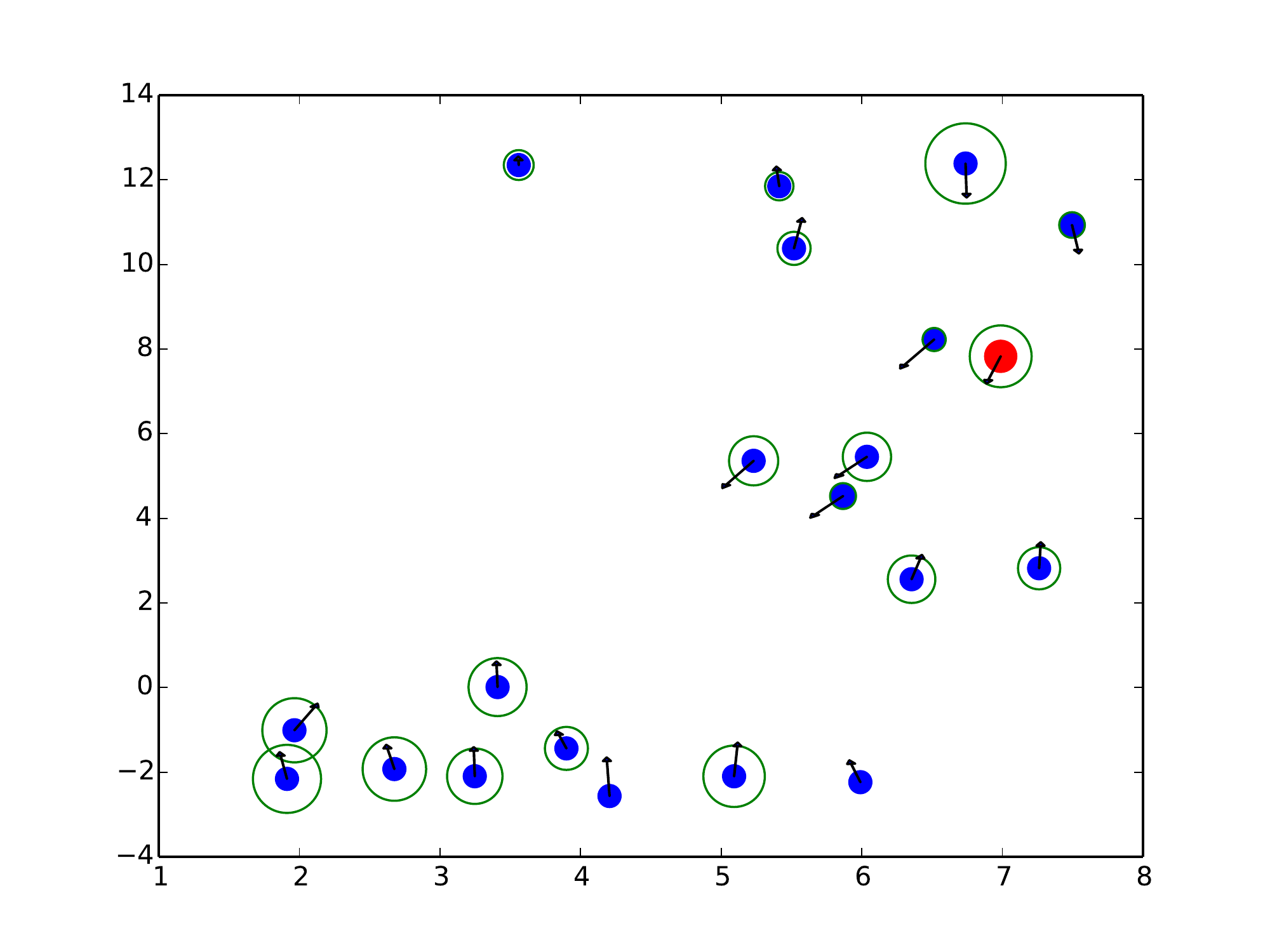}&%
		\includegraphics[width=0.24\linewidth]{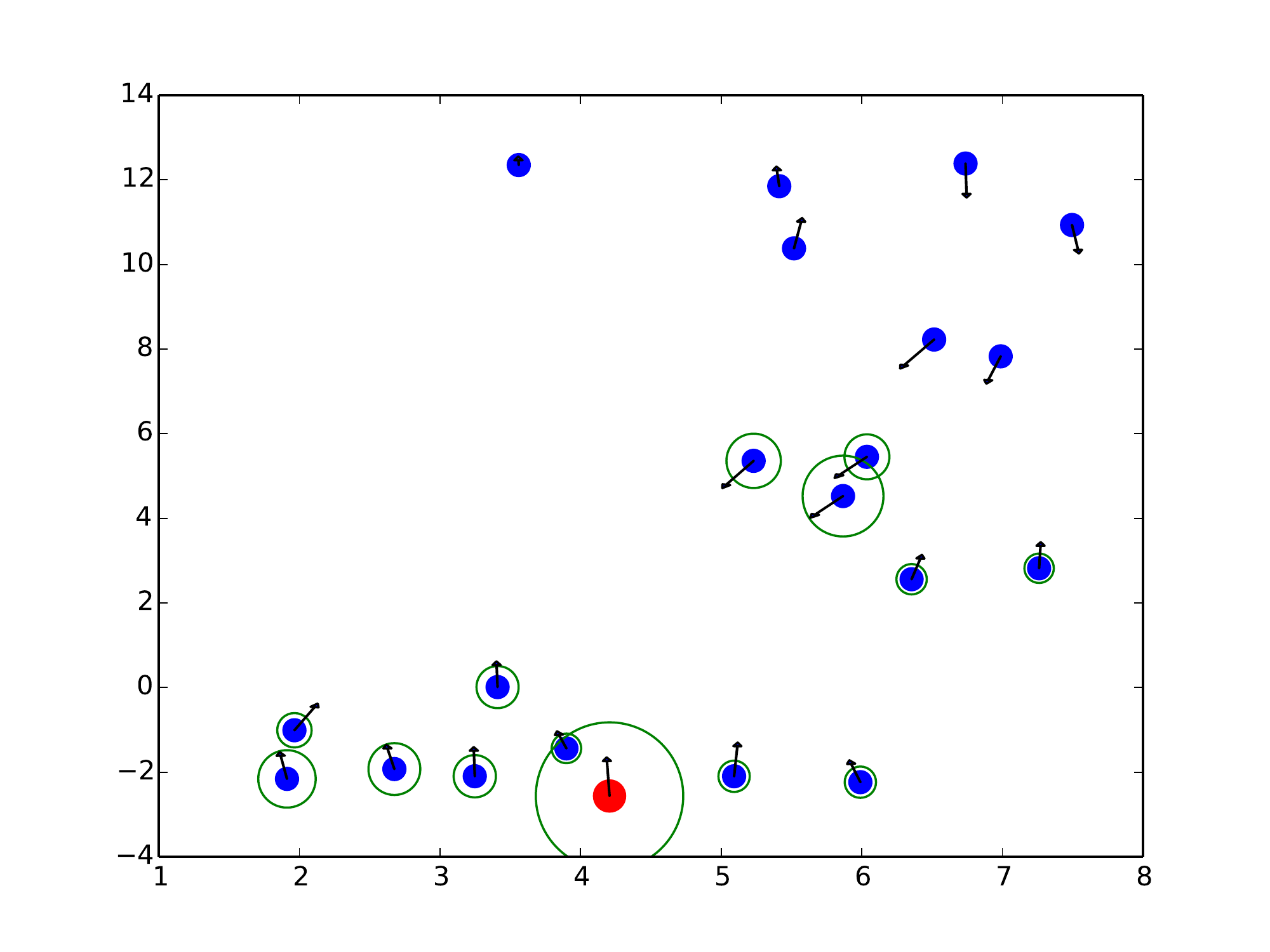}&%
		\includegraphics[width=0.24\linewidth]{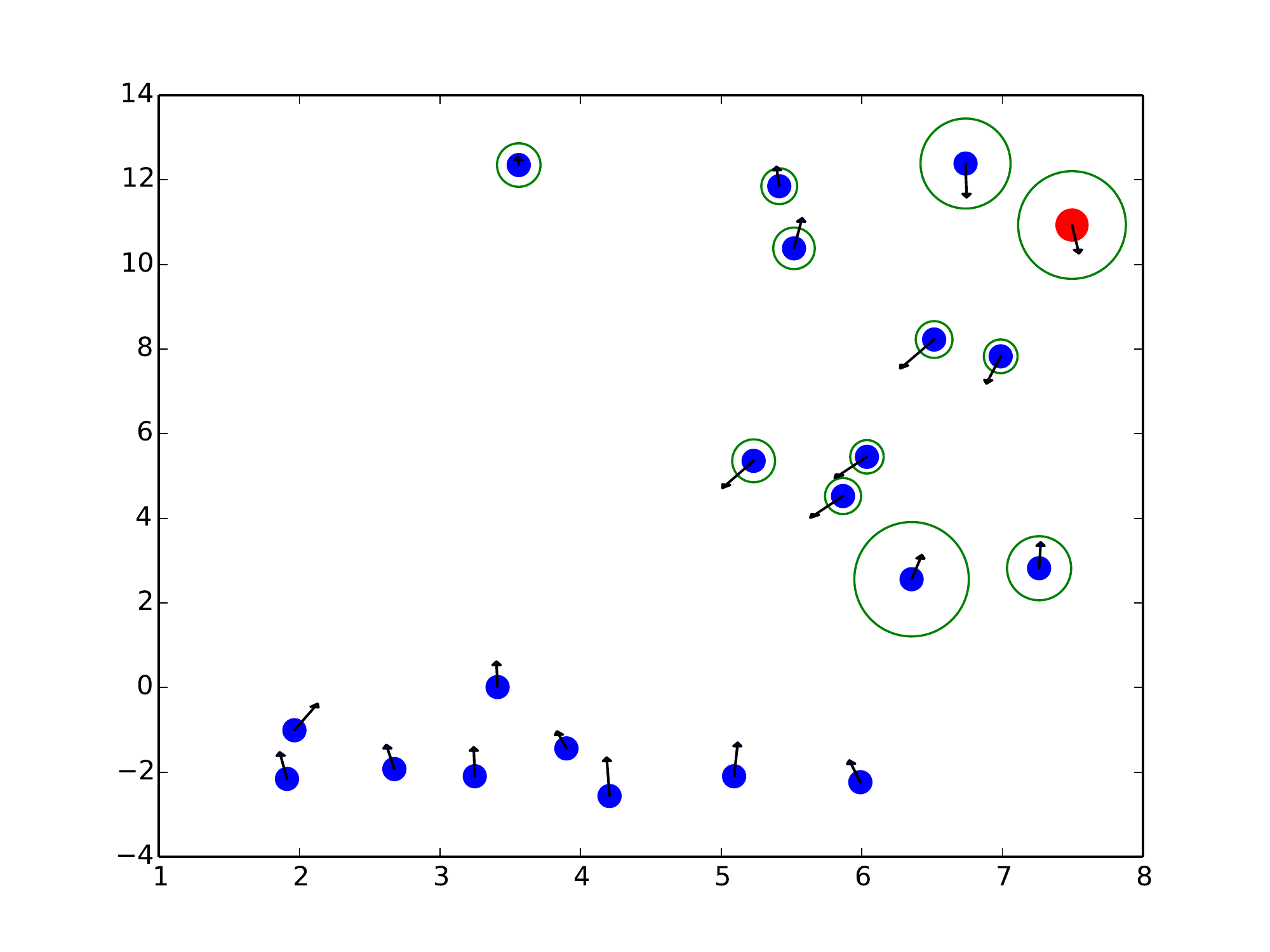}&%
		\includegraphics[width=0.24\linewidth]{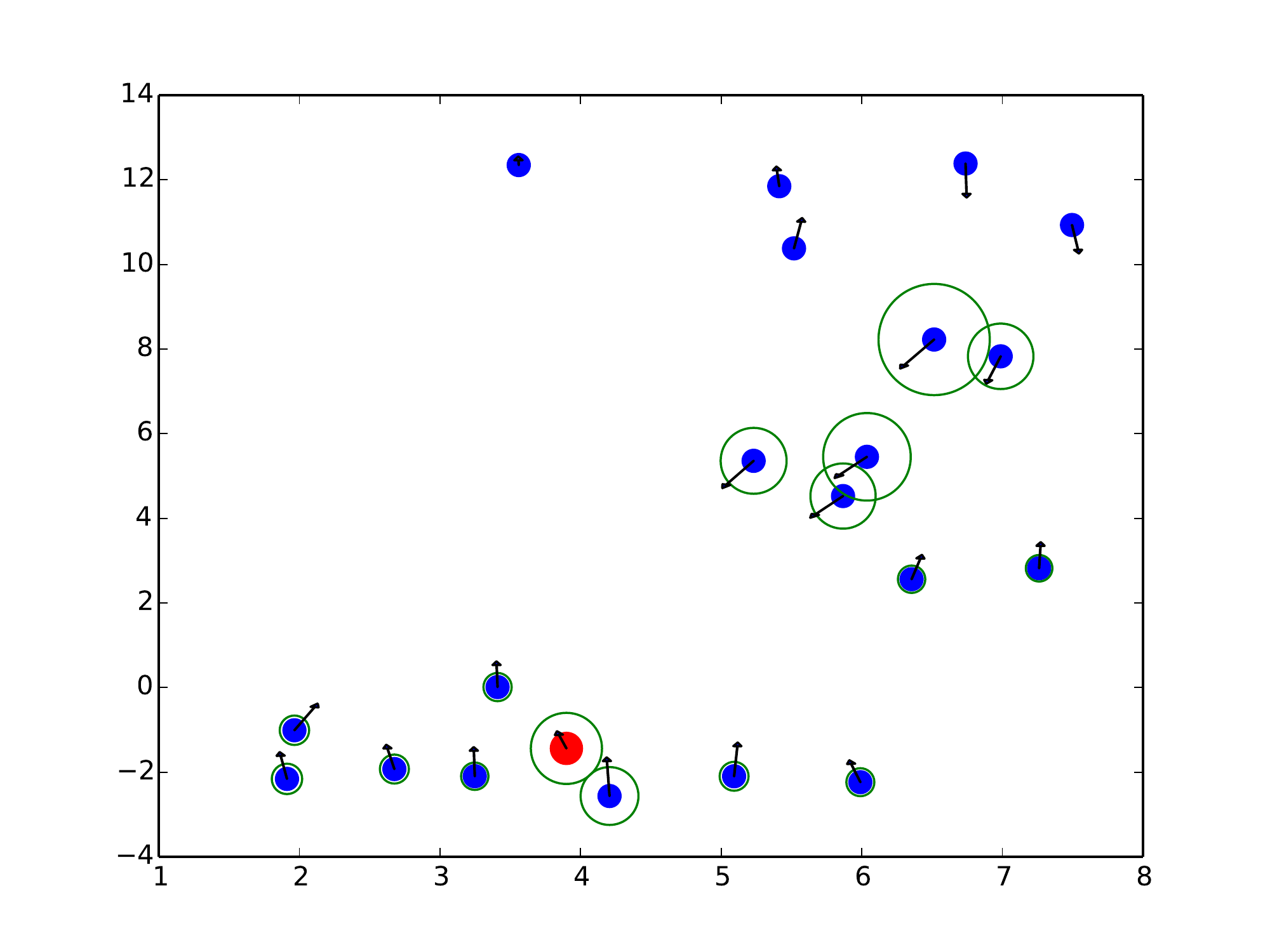}%
	\end{tabular}
	\centering
	\caption{\small Additional attention visualizations of the spatial Transformer, i.e., connected in the interaction graph, in encoder 2. We visualize the attention of neighbor pedestrians with respect to the red dotted pedestrian. The size of circles represents the attention value and bigger circles indicate higher attention. STAR learns reasonable spatial attention, the pedestrians have higher attentions over themselves and their neighbors. }
	\label{fig:att}
\end{figure*}

\clearpage
\section*{Ablation Trajectory Prediction Visualizations}
\begin{figure*}[!htb]
	\fontsize{6}{11}\selectfont
	\centering
	\begin{tabular}{c c c c}
		\includegraphics[width=0.24\linewidth]{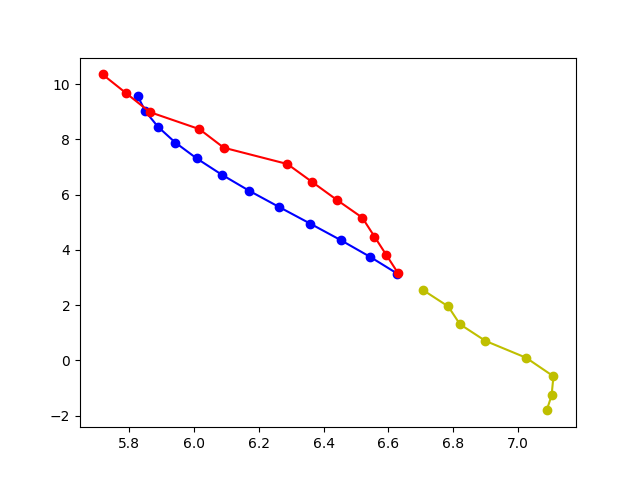}&%
		\includegraphics[width=0.24\linewidth]{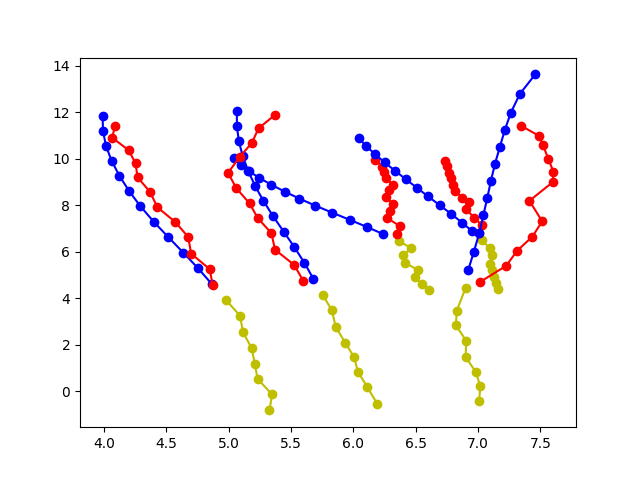}&%
		\includegraphics[width=0.24\linewidth]{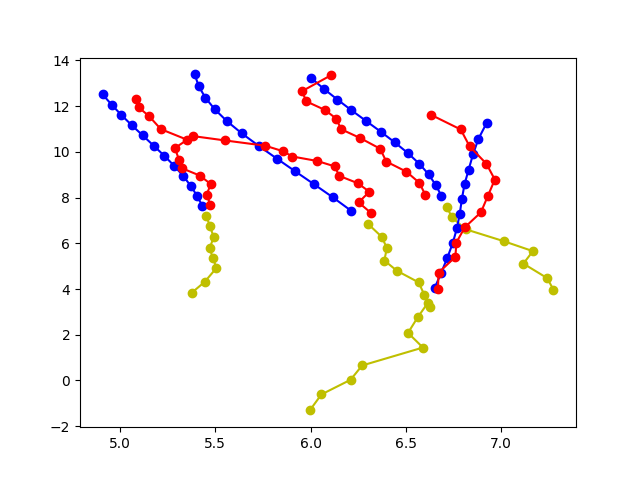}&%
		\includegraphics[width=0.24\linewidth]{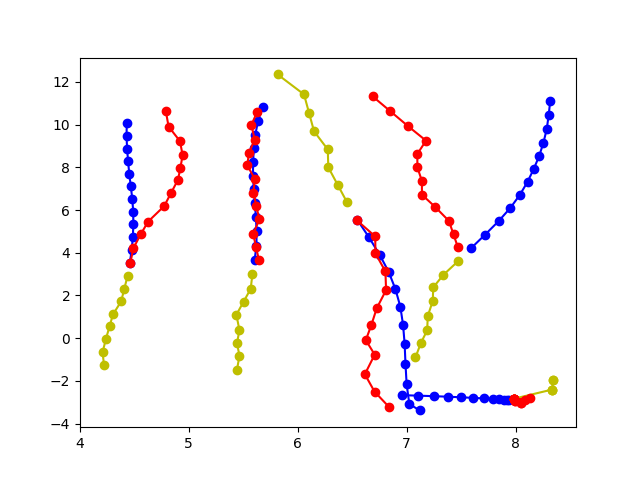} \\
		\multicolumn{4}{c}{(a) GAT + STAR}\\
		\includegraphics[width=0.24\linewidth]{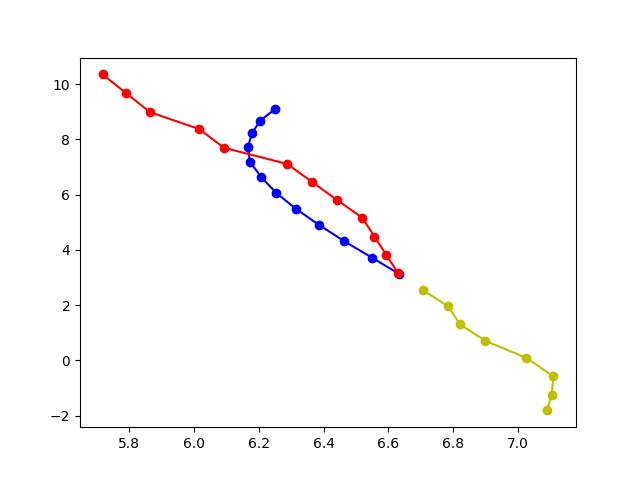}&%
		\includegraphics[width=0.24\linewidth]{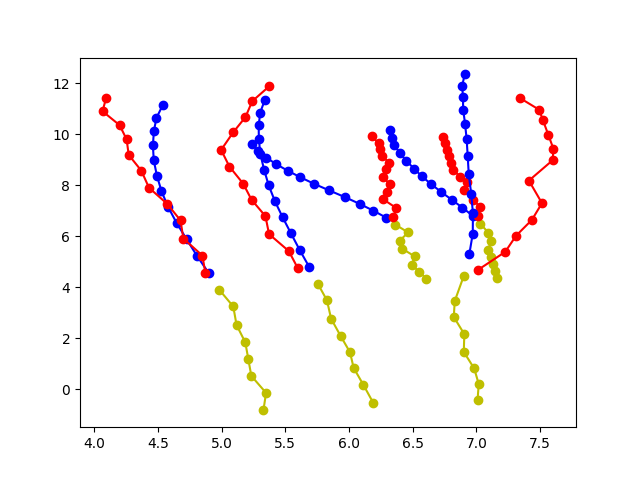}&%
		\includegraphics[width=0.24\linewidth]{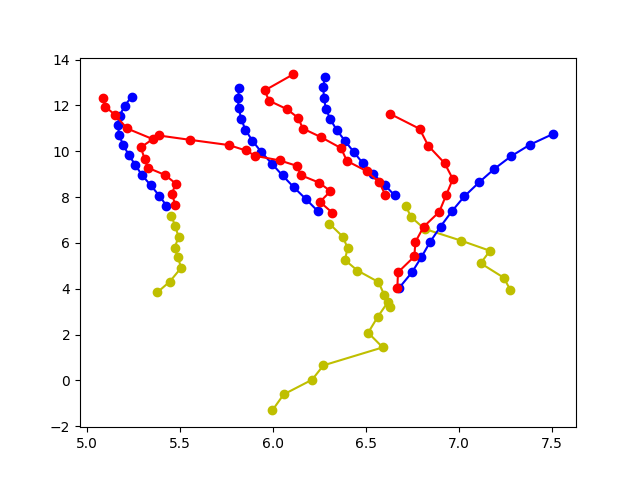}&%
		\includegraphics[width=0.24\linewidth]{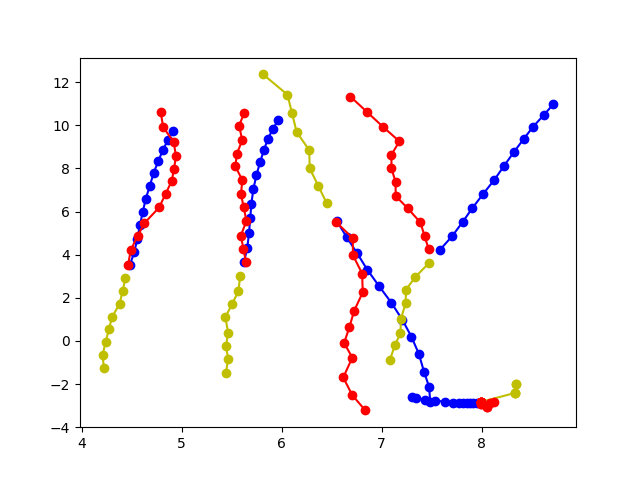} \\
		\multicolumn{4}{c}{(b) MHA + STAR}\\
		\includegraphics[width=0.24\linewidth]{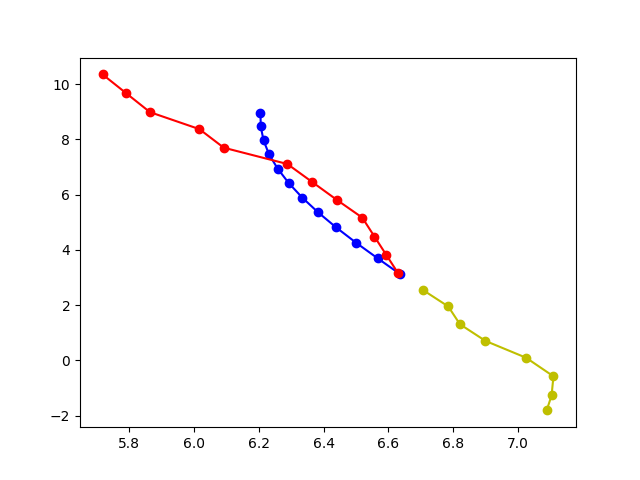}&%
		\includegraphics[width=0.24\linewidth]{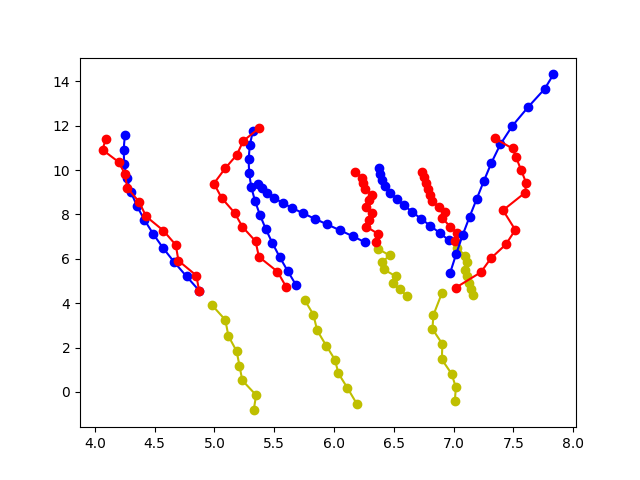}&%
		\includegraphics[width=0.24\linewidth]{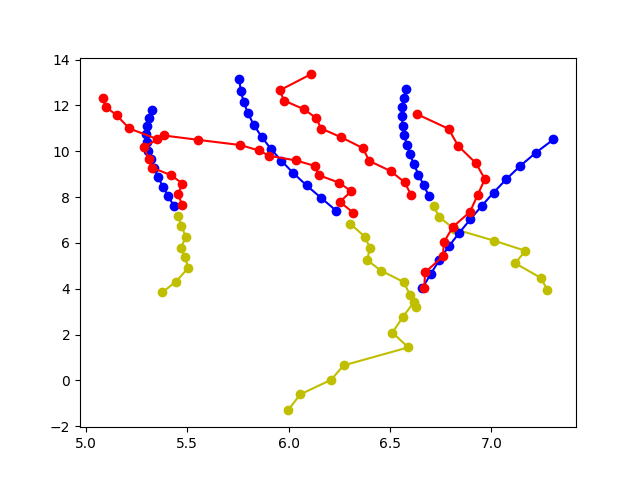}&%
		\includegraphics[width=0.24\linewidth]{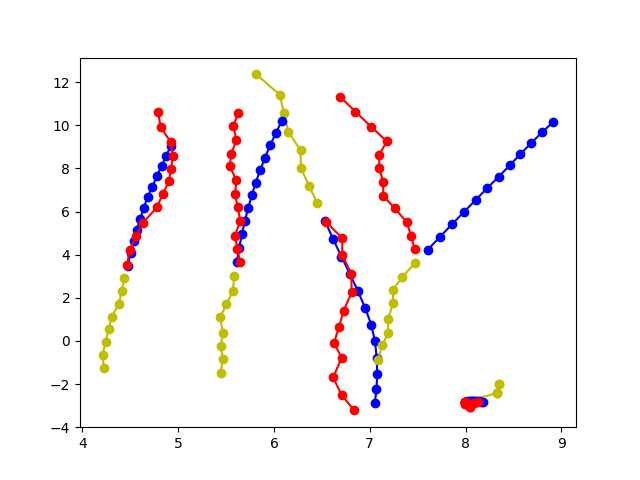} \\
		\multicolumn{4}{c}{(c) STAR + LSTM}\\
		\includegraphics[width=0.24\linewidth]{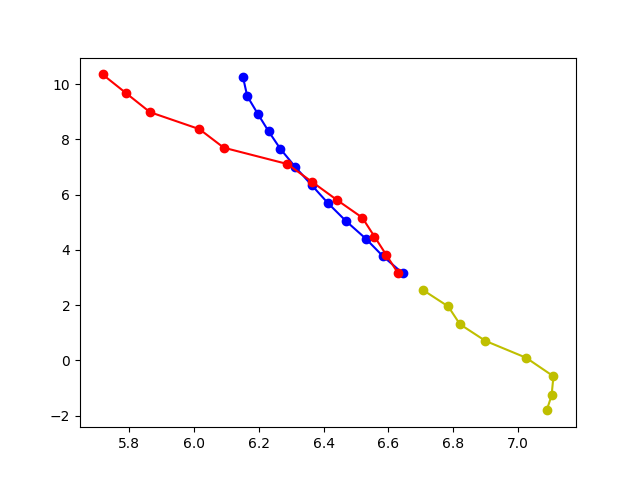}&%
		\includegraphics[width=0.24\linewidth]{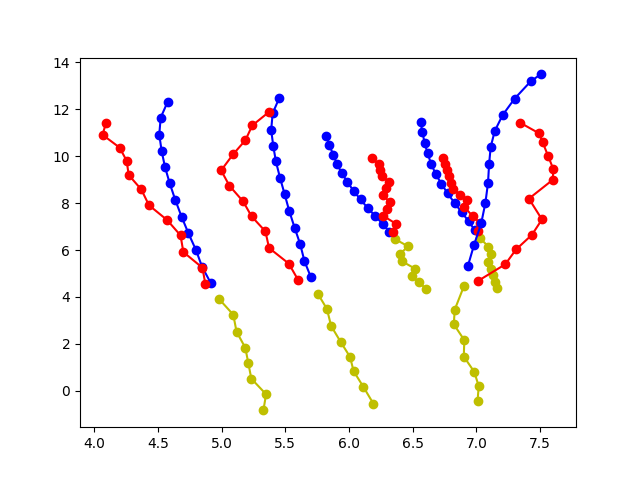}&%
		\includegraphics[width=0.24\linewidth]{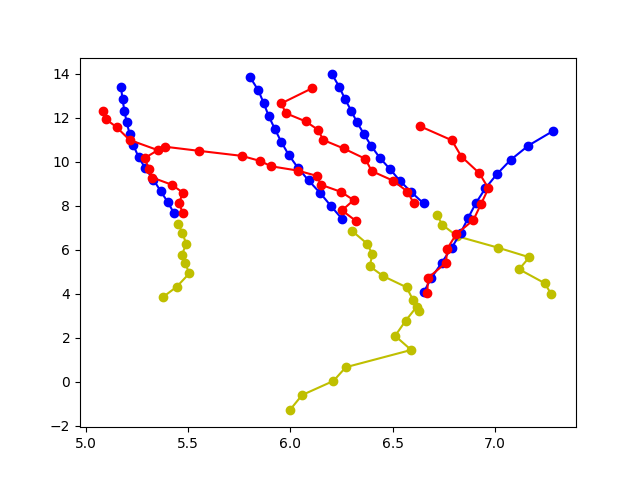}&%
		\includegraphics[width=0.24\linewidth]{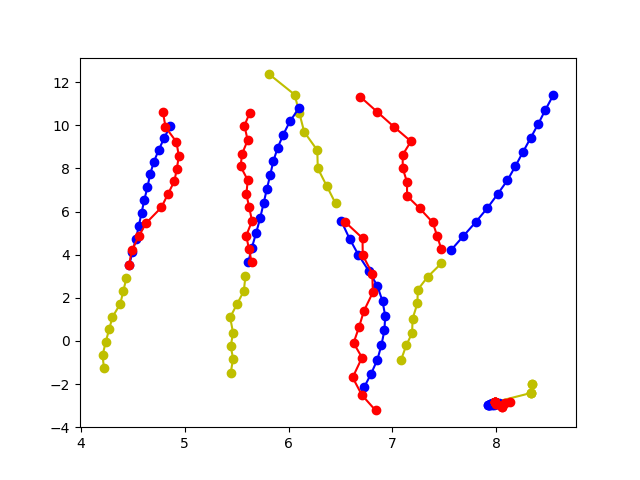} \\
		\multicolumn{4}{c}{(d) STAR without Graph Memory}\\
		\includegraphics[width=0.24\linewidth]{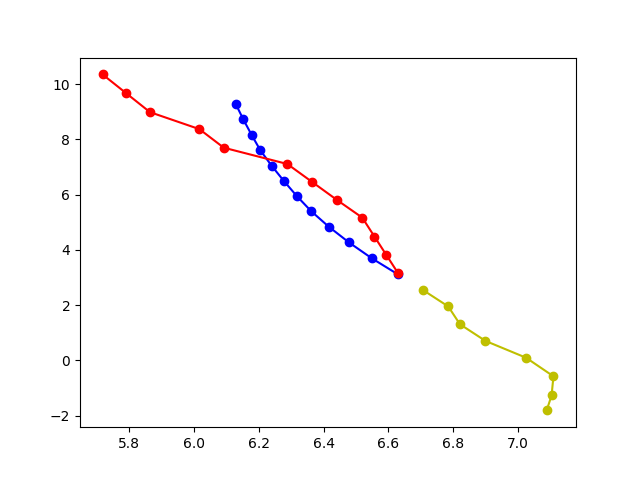}&%
		\includegraphics[width=0.24\linewidth]{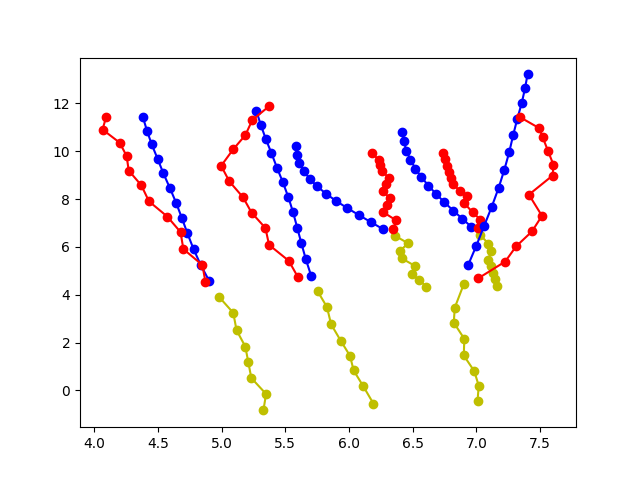}&%
		\includegraphics[width=0.24\linewidth]{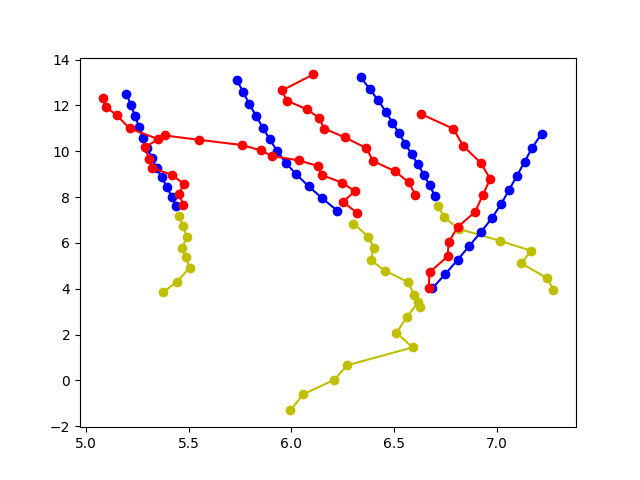}&%
		\includegraphics[width=0.24\linewidth]{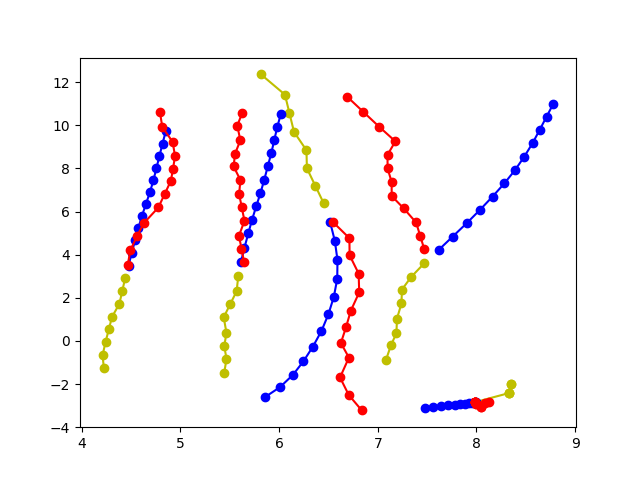} \\
		\multicolumn{4}{c}{(e) Simplified STAR without Encoder 2}\\
		\includegraphics[width=0.24\linewidth]{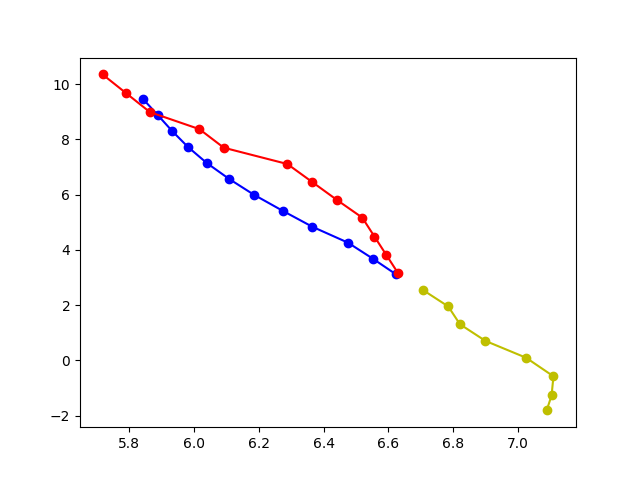}&%
		\includegraphics[width=0.24\linewidth]{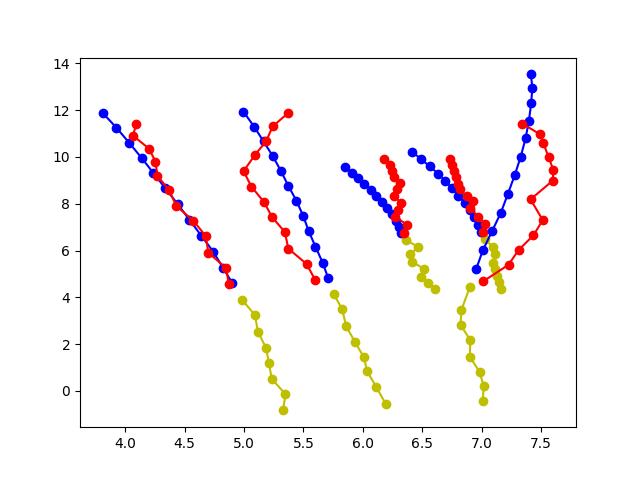}&%
		\includegraphics[width=0.24\linewidth]{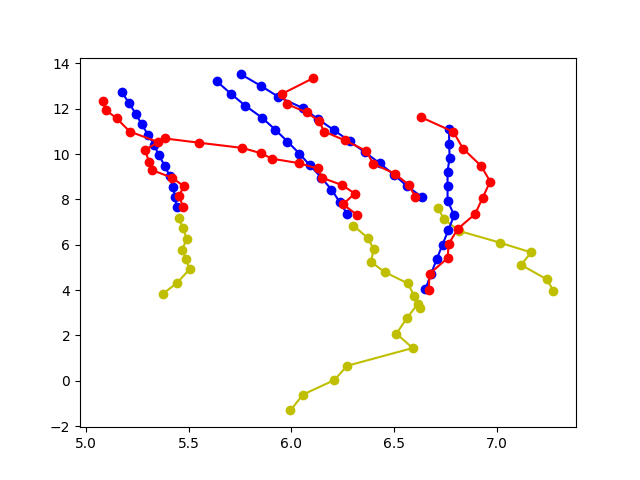}&%
		\includegraphics[width=0.24\linewidth]{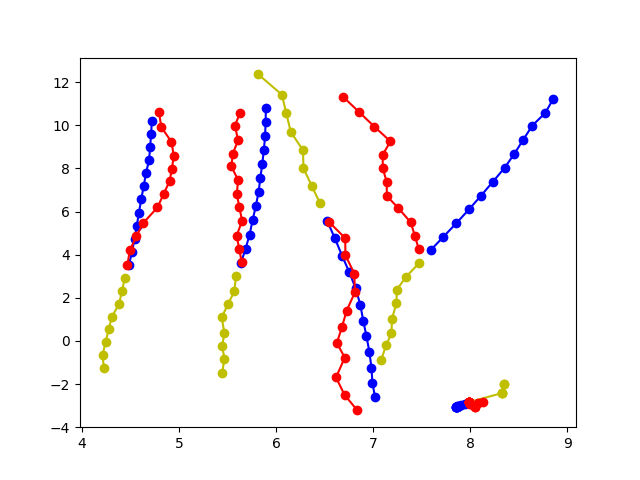} \\
		\multicolumn{4}{c}{(f) STAR}\\
	\end{tabular}
	\centering
	\caption{\small Trajectory visualization of all ablations. Yellow lines denote the history, red lines denote the ground-truth, and blue lines denote the prediction. Qualitatively, STAR produces best predictions both spatially and temporally.}
	\label{fig:pred}
\end{figure*}
\end{document}